\documentclass[runningheads]{llncs}

\usepackage{accv}

\usepackage{accvabbrv}
\usepackage{todonotes}
\usepackage{graphicx}
\usepackage{booktabs}
\usepackage{bm}
\usepackage{multirow}
\usepackage{color, colortbl}
\usepackage{textcomp}
\usepackage[export]{adjustbox}
\newcommand{\croppdf}[1]{\IfFileExists{#1-crop.pdf}{\immediate\write18{pdfcrop #1.pdf}}{\immediate\write18{pdfcrop #1.pdf}}}
\definecolor{LightCyan}{rgb}{0.88,1,1}
\newcommand{\subbest}[1]{\textbf{{#1}}}
\newcommand{\subsecond}[1]{\underline{{#1}}}

\newcommand{\Paragraph}[1]{\textbf{#1}}
\usepackage[accsupp]{axessibility}  % Improves PDF readability for those with disabilities.
\newcommand{\mybm}[1]{\mathbf{#1}}

\usepackage[pagebackref,breaklinks,colorlinks,citecolor=accvblue]{hyperref}
\usepackage{orcidlink}

\begin{document}


\title{LoG-VMamba \includegraphics[width=0.68cm]{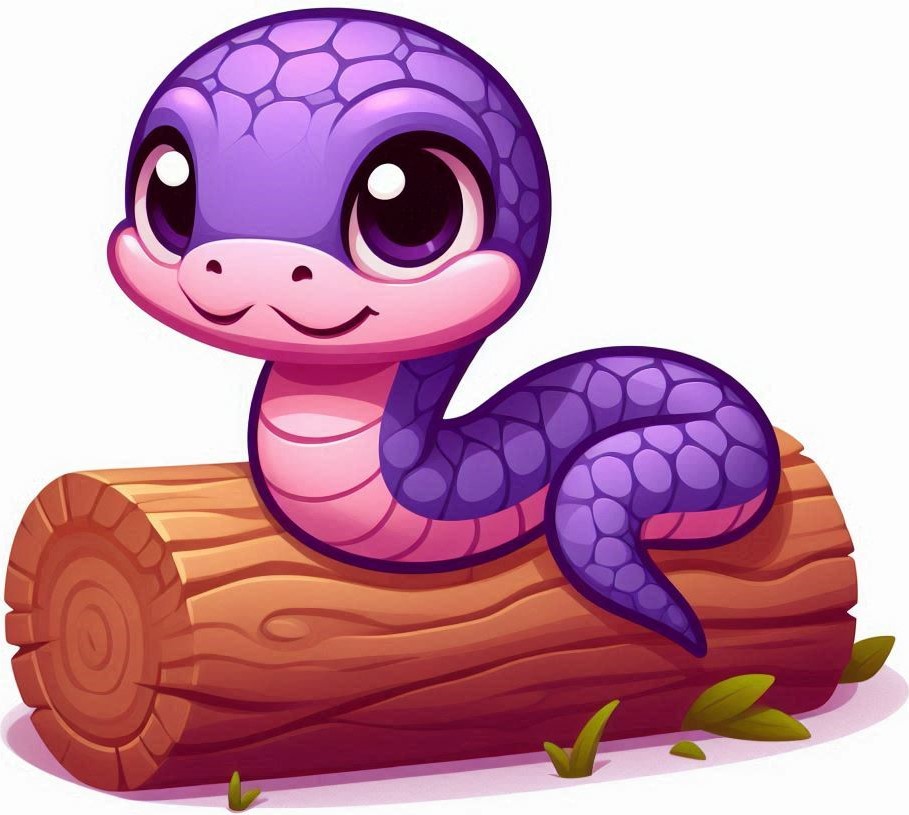}: Local-Global Vision Mamba \\ for Medical Image Segmentation} 
\titlerunning{LoG-VMamba: Local-Global Vision Mamba}

\author{Trung DQ. Dang
\and
Huy Hoang Nguyen \and
Aleksei Tiulpin}
%index{Dang, Trung}
%index{Nguyen, Huy Hoang}
%index{Tiulpin, Aleksei}

\authorrunning{T. Dang et al.}

\institute{University of Oulu, Finland \\
\email{\{trung.ng,huy.nguyen,aleksei.tiulpin\}@oulu.fi}}

\maketitle

% \begin{figure}[t]
%     \centering
%     \croppdf{figures/flops_perf_endoscopy}
%     \croppdf{figures/flops_perf_cell}
%     % \hspace*{\fill}
%     \subfloat[Endoscopy dataset]{\includegraphics[width=0.45\textwidth]{figures/flops_perf_endoscopy-crop.pdf}}
%     \hfill
%     \subfloat[Cell dataset]{\includegraphics[width=0.45\textwidth]{figures/flops_perf_cell-crop.pdf}}
%     % \hspace*{\fill}
%     \caption{Computional efficiency and performance comparisons between our method }
%     \label{fig:enter-label}
% \end{figure}

%===========================================================
\begin{abstract}
Mamba, a State Space Model (SSM), has recently shown competitive performance to Convolutional Neural Networks (CNNs) and Transformers in Natural Language Processing and general sequence modeling. Various attempts have been made to adapt Mamba to Computer Vision tasks, including medical image segmentation (MIS). Vision Mamba (VM)-based networks are particularly attractive due to their ability to achieve global receptive fields, similar to Vision Transformers, while also maintaining linear complexity in the number of tokens. However, the existing VM models still struggle to maintain both spatially local and global dependencies of tokens in high dimensional arrays due to their sequential nature. Employing multiple and/or complicated scanning strategies is computationally costly, which hinders applications of SSMs to high-dimensional 2D and 3D images that are common in MIS problems. In this work, we propose Local-Global Vision Mamba, LoG-VMamba, that explicitly enforces spatially adjacent tokens to remain nearby on the channel axis, and retains the global context in a compressed form. Our method allows the SSMs to access the local and global contexts even before reaching the last token while requiring only a simple scanning strategy. Our segmentation models are computationally efficient and substantially outperform both CNN and Transformers-based baselines on a diverse set of 2D and 3D MIS tasks. The implementation of LoG-VMamba is available at \url{https://github.com/Oulu-IMEDS/LoG-VMamba}.

\end{abstract}

%===========================================================
\section{Introduction}

Medical image segmentation (MIS) targets the delineation and location of tissues and lesions in 2D or 3D medical images. This process is crucial for developing automatic disease identification, staging, and treatment, as well as developing medical robotics. In recent years, the state-of-the-art approaches to MIS have been based on deep learning (DL), thanks to its ability to learn representations of complex patterns from large datasets. This proves to be essential in producing production-quality performances in medical applications ~\cite{ronneberger2015u,zhou2019unet++,hatamizadeh2022unetr,cao2022swin,hatamizadeh2021swin,xiao2018weighted}.

% Several prior works have proposed using Vision Mamba models in medical image segmentation, and we follow in their footsteps. We plug our proposed Vision Mamba modules into a few previous Vision Mamba segmentation models and conduct extensive experiments on both 2D and 3D datasets. The experimental results on 4 datasets demonstrate that our modules achieves superior performance, surpassing the traditional Vision Mamba modules by a significant margin.

Visual feature extraction plays a vital role in Computer Vision (CV) tasks, including MIS. In the early stage of DL, convolutional neural network (CNN) was the ubiquitous feature extraction module, since this architecture is effective in learning local patterns by taking into account the regional context around pixels~\cite{he2016deep,huang2017densely,liu2022convnet}. However, one drawback of CNN is the difficulty in capturing long-range dependencies (LRD), which is essential for extracting high-level features from data~\cite{knigge2023modelling}. To enable LRD capturing (i.e. increasing the receptive field), some studies stacked a large number of layers with downsampling~\cite{lecun2015deep}, while others used dilated convolutions~\cite{yu2015multi}. However, the former is computationally expensive and the latter may result in missing fine-grained details. 
% \todo{Here we need to have a stronger closing statement. The previous one was not strong. Back up by the reference.}

In contrast to CNNs, Vision Transformer (ViT)-based architectures rely on the attention mechanism to learn LRD across the entire image~\cite{dosovitskiy2020image,zhou2021deepvit}. Despite their global receptive field (GRF) at all levels of learned representations, ViTs are hindered by their quadratic complexity with respect to the number of tokens. In the case of ViTs, these tokens are the patches into which an image is split before it is fed into the network. Therefore, utilizing ViTs in tasks with high-dimensional inputs and outputs such as semantic segmentation may create computational efficiency issues~\cite{liu2024vmamba}.

\begin{figure}[t]
  \centering
  \croppdf{figures/mamba_issue_sa}
  \croppdf{figures/mamba_issue_conv}
  \croppdf{figures/mamba_issue_ss2d_h}
  \croppdf{figures/mamba_issue_ss2d_v}
  \subfloat[Convolution\label{fig:conv_locality}]{\includegraphics[width=0.23\textwidth]{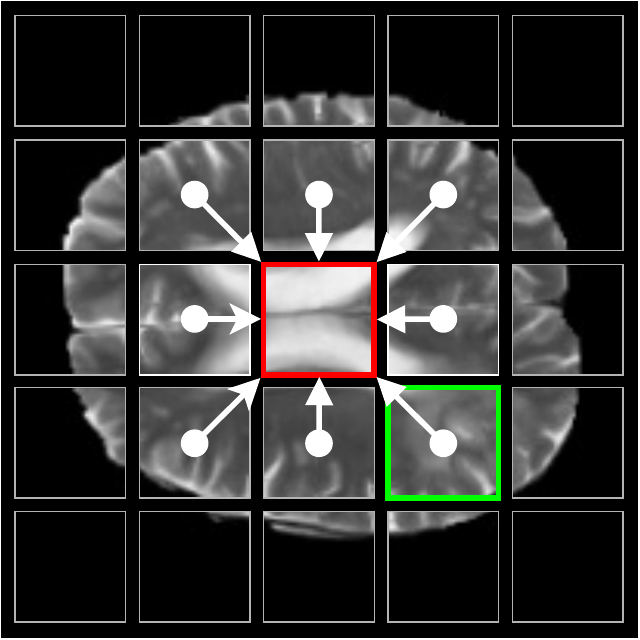}}
  \hfill
  \subfloat[Self-attention\label{fig:transformer_locality}]{\includegraphics[width=0.23\textwidth]{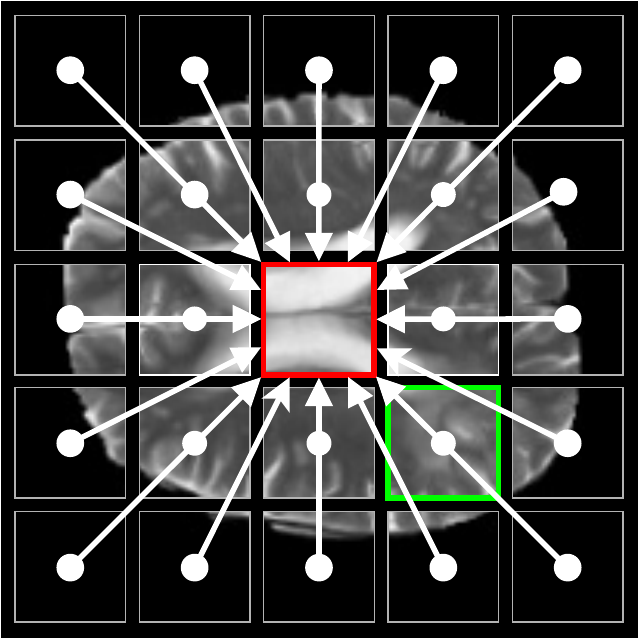}}
  \hfill
  \subfloat[Horizontal scan\label{fig:ss2d_h}]{\includegraphics[width=0.23\textwidth]{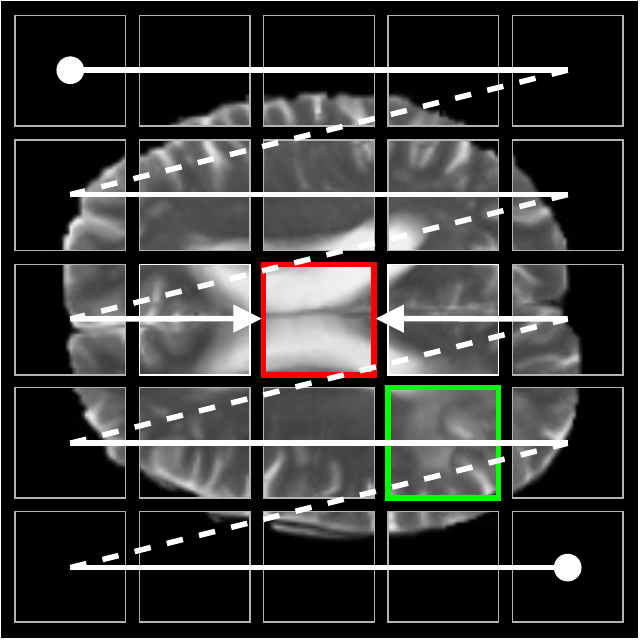}}
  \hfill
  \subfloat[Vertical scan\label{fig:ss2d_v}]{\includegraphics[width=0.23\textwidth]{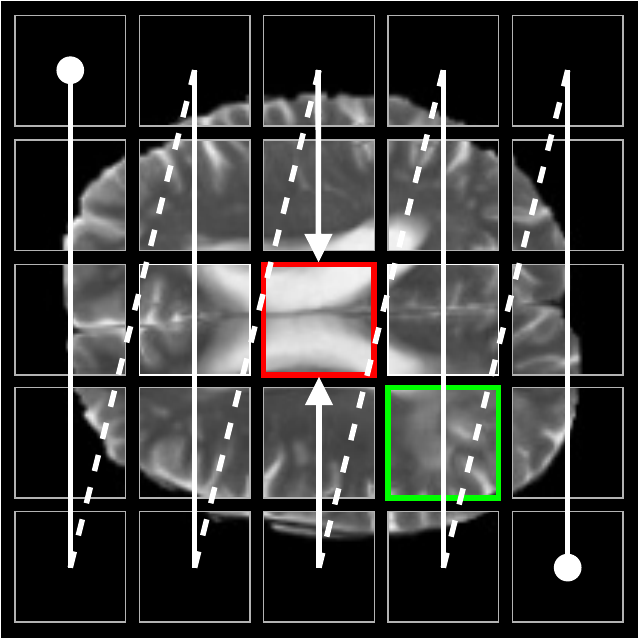}}
  \caption{The comparison between how feature extractors establish a correlation between the query token (in red) and a neighboring token (in green). 
  % The red boxes indicate the query tokens, while the purple boxes indicate one of the query's neighboring tokens. 
  In (c-d)~\cite{liu2024vmamba}, the distance between the query and its neighbor may be roughly one row (or column) of tokens.}
  \label{fig:mamba_issue}
\end{figure}

Recently, Mamba~\cite{gu2023mamba}, a State Space Model (SSM)-based method, has been introduced in the field of Natural Language Processing (NLP). Mamba enables the acquisition of GRF via its recurrent mechanism,
% tokenizing the input data into a sequence of tokens similar to ViTs,
% \todo{Tokenizing the input has nothing to do with GRF. What leads to GRF is either self-attention or recurrent mechanism. Tokens are just a representation that even CNN can run on, but without GRF.}
while maintaining computational efficiency due to the linearity of complexity w.r.t. data dimension. Several studies have been conducted to adapt Mamba to CV tasks such as image classification and MIS~\cite{zhu2024vision,ma2024u,liu2024swin,xing2024segmamba}. Despite the computational benefits, in vision applications~\cite{liu2024vmamba}, Mamba struggles to maintain dependencies between both neighboring and distant tokens due to the restricted capacity of Mamba's state~\cite{shi2024multi} and the necessity of a sequential approach to model LRD.
% achieving Mamba's GRF.
% \todo{Unclear. Regardless of scanning strategy, dependencies between both neighboring and distant pixels are possible due to the state capturing all information of previous states. What makes Mamba fail is the limited size of the state, which makes compressing all information difficult.}. 
\cref{fig:mamba_issue} demonstrates the weakness of two common scanning protocols of Vision Mamba (VM)-based methods~\cite{liu2024vmamba,liu2024swin,xing2024segmamba} compared to convolutions and ViTs. Some studies attempted to maintain the locality of neighboring tokens using a complicated scanning strategy such as a multidirectional or omnidirectional scan~\cite{xing2024segmamba,zhao2024rs}. However, these significantly increase the computational complexity.

In this work, we argue that developing more complex scanning strategies is unnecessary, should there be an image-friendly tokenizer. As such, we propose the \textbf{Local-Global Vision Mamba (LoG-VMamba)}, whose core components are Local Token eXtractor (LTX) and Global Token eXtractor (GTX). 
LTX maintains the locality of neighboring tokens in high-dimensional arrays, explicitly ensuring that spatially nearby tokens remain close along the channel axis. GTX, on the other hand, compresses features across all spatial dimensions, providing the SSM module with compressed versions of GRF before reaching the final time step.
The combined effects of the two components make LoG-VMamba significantly distinct from prior VM-based methods as we eliminate the need for a sophisticated scanning strategy. In summary, our study has the following contributions:
% which are designed to tackle the tokenization issues in CV applications, and in particular, MIS. Our contributions are:
% \todo[inline]{Contrib. 2 does not see to add value}
\begin{enumerate}
    \item We propose two visual token extractors, LTX and GTX, that provide the SSMs with both local and global visual contexts at early time steps, inspired by the strengths of both CNN and ViT.
    \item We propose LoG-VMamba, a Mamba-based module for CV, leveraging the power of both LTX and GTX. Based on LoG-VMamba, we introduce segmentation models for 2D and 3D MIS problems.
    % \todo{Perhaps we can add segmentation models here too.}
    % We propose a method \todo{two methods instead of a method. Aleksei's comment: no, we propose one method that has two components. A paper cannot tell more than 1 story. The number of stories is only one. Logvmamba is a method. } that allows Mama handle both local and global information at a token level through the GTX and LTX modules.
    % \item HERE SHOULD BE A SECOND CONTRIBUTION. \todo{Put LTX and GTX in 2 lines. They are independent.}
    \item Our experiments demonstrate that the developed models consistently outperform multiple well-known baselines in a variety of 2D and 3D MIS benchmarks. We also show that our approach does not require an advanced scanning strategy to achieve state-of-the-art performance, and is thus computationally efficient.
\end{enumerate}%While LTX aims to maintain the locality of neighboring tokens in high-dimensional arrays, GTX is designed to provide the SSM module with a GRF. With these modules, LoG-VMamba becomes an SSM-based network that possesses the strengths of both CNN and ViT. Specifically, LTX has an unfolding operator that allows us to explicitly enforce spatially nearby tokens to stay closely along the channel axis.
\section{Related Work}

\Paragraph{Feature Extraction.} The evolution of feature extraction modules has been central to the field of CV. After the success of classic CNNs such as AlexNet~\cite{krizhevsky2017imagenet} and VGG~\cite{simonyan2014very}, learnable feature extractors have gained widespread adoption and a plethora of improvements to CNN design have been developed~\cite{he2016deep,hu2018squeeze,chollet2017xception,howard2017mobilenets,yu2015multi}. These works have improved the performance of CNNs on various CV tasks, and they showcased the capabilities of CNNs to learn local patterns at low-level layers, as well as hierarchical representations at high-level layers. Afterward, inspired by the success of Transformers~\cite{vaswani2017attention} in sequence modeling, ViT~\cite{dosovitskiy2020image} emerged and challenged the dominance of CNNs in vision-related problems. ViT tokenizes an image along spatial dimensions into patches and employs multi-head self-attention (MSA) over the obtained sequence of tokens. Thanks to MSA, ViT is able to handle LRD at the early layers of a vision model. Numerous subsequent works~\cite{dong2022cswin,hassani2023neighborhood,liu2021swin,yu2022metaformer,touvron2021training} enhanced various aspects of ViT, such as efficiency and refined attention mechanisms. Moreover, some studies~\cite{dai2021coatnet,guo2022cmt,lin2023scale,pan2022integration} aimed to combine the desirable properties of both CNNs and ViTs.

\Paragraph{State Space Models.} Inspired by the principles from control theory, SSMs~\cite{gu2023mamba,gu2021efficiently,gu2020hippo} have emerged as an alternative to recurrent neural networks~\cite{orvieto2023resurrecting} and Transformers~\cite{vaswani2017attention} in NLP and sequence modeling. 
% The classical continuous SSMs map an input sequence to a latent state and then predict an output sequence. 
To adapt these models into DL, prior studies~\cite{gu2023mamba,gu2021efficiently} employed the zero-order hold with a timescale parameter to transform the operations of SSMs into discrete form. Mamba~\cite{gu2023mamba} let these parameters of SSMs be input-dependent and designed a hardware-aware algorithm to boost the throughput. A plethora of works~\cite{liu2024vmamba,zhu2024vision,wang2024mamba,huang2024localmamba,shi2024multi,zhao2024rs} explored the viability of SSMs in visual feature extraction. Most of these~\cite{liu2024vmamba,zhu2024vision,huang2024localmamba,zhao2024rs} focused on designing suitable scanning strategies for processing images. Instead of finding how to \textit{scan} the tokens, we focus on how to \textit{build} informative visual tokens embedded with local-global dependencies. 

\Paragraph{Medical Image Segmentation.} Progress in segmentation, particularly MIS, has been built upon the advancements in representation learning. U-Net~\cite{ronneberger2015u} is the most prominent early representative of a U-shaped architecture, which influenced the design of nearly all modern MIS approaches, and it consists of an encoder and a decoder connected at each feature scale via skip connections. While the traditional U-Net uses pooling and transposed convolution layers to change the feature scale, its pure Transformer variant~\cite{cao2022swin} replaces them with patch merging and patch expanding layers. Afterward, the trend of following this architecture~\cite{xiao2018weighted,li2018h,hatamizadeh2022unetr,hatamizadeh2021swin} has persisted in MIS until 
the recent emergence of Mamba-based models such as U-Mamba~\cite{ma2024u} and Swin-UMamba~\cite{liu2024swin}. These Mamba-inspired architectures have demonstrated advances in both computational efficiency and effectiveness when compared to CNN and Transformer-based methods in MIS.
% Liu~\textit{et al.}~\cite{liu2024swin} presented Swin-UMamba along with its variant Swin-UMamba$^\dagger$. The decoder in the former model utilizes CNN, whereas the latter version incorporates visual state space blocks in its decoder. Ma~\textit{et al.} proposed U-Mamba 
In this study, we further enhance the performance of these two methods using our LoG-VMamba module in 2D and 3D MIS.

% \todo{I would still bring the attention of the reader to the importance of long-range dependencies in MIS. SSM paragraph further seems to be disconnected from this one. We need to talk about tokens, sequences, and why Transformers are not good for LRD.}

%===========================================================
\section{Methodology}

% \begin{figure}[tb]
%   \centering
%   \croppdf{figures/dc_mamba}
%   \includegraphics[width=\textwidth]{figures/dc_mamba-crop.pdf}
%   \caption{The overview of Global-local Vision Mamba block}
%   \label{fig:dc_mamba}
% \end{figure}

% \todo{Feels like there is a lot of empty space in Figure 2.}
% \todo{Is Figure 1 the main message? Also, it reads in Figure 3 you have all tokens combined being HxWxC. I guess it is not the case? You probably have N tokens, innit? }
\subsection{Preliminaries}

% We assume a 2D segmentation scenario in this section for simplicity, but our formulation can be straightforwardly extended to a 3D segmentation setting. Given a 2D input image of size $H \times W \times C$, a semantic segmentation model is expected to produce an output of size $H \times W \times K$, where $K$ is the number of classes. A typical semantic segmentation model is composed of an encoder and a decoder. The encoder extracts relevant features from the image, where each feature map $x_t$ is of size $H_i \times W_i \times C_i$. Meanwhile, the decoder aims to fuse features from different levels of scale and eventually predicts a segmentation map.

\Paragraph{State Space Model}. Time-continuous SSMs~\cite{kalman1960new} refer to linear-time-invariant systems formulated as follows
\begin{align}
    \mybm{h}'_t &= \mybm{A}\mybm{h}_t + \mybm{B}\mybm{x}_t \\
    \mybm{y}_t &= \mybm{C}\mybm{h}_t,
\end{align}
where $\mybm{x}_t$ and $\mybm{y}_t$ are $D$-dimensional input and output vectors, respectively. $\mybm{A}$, $\mybm{B}$, and $\mybm{C}$ are learnable parameters independent of $\mybm{x}_t$.

% \Paragraph{Structured State Space Model} Let $\{\mybm{x}_t\}_{t\in [L]}$ represent an input sequence of row vectors $\mybm{x}_t \in \mathbb{R}^{1 \times D}$, where $[L] = \{1,\dots,L\}$. Structured SSM (S4)~\cite{gu2021efficiently} is formulated as
% \begin{align}
%     \mybm{h}'_t &= \mybm{A}\mybm{h}_t + \mybm{B}\mybm{x}_t \\
%     \mybm{y}_t &= \mybm{C}\mybm{h}_t,
% \end{align}
% where $t\in[L]$, $\mybm{h}_t$ is a hidden state and $\mybm{y}_t$ is the resulting row vector, while $\mybm{A}$, $\mybm{B}$, and $\mybm{C}$ are learnable parameters independent of $\mybm{x}_t$.

\Paragraph{Selective State Space Model}. Selective SSM (S6)~\cite{gu2023mamba} consists of a hidden state $\mybm{h}_t$, three continuous parameters $\mybm{A}$, $\mybm{B}$, and $\mybm{C}$ as well as a step size parameter $\Delta$. Among these learnable parameters, $\mybm{B}$, $\mybm{C}$, and $\Delta$ depend on $\mybm{x}_t$, making S6 distinct from prior SSMs. Let $\{\mybm{x}_t\}_{t\in [L]}$ represent an input sequence of row vectors $\mybm{x}_t \in \mathbb{R}^{1 \times D}$, where $[L] = \{1,\dots,L\}$.
% Specifically, at time step $t$, $\mybm{B}_t$, $\mybm{C}_t$, and $\myDelta_t$ are computed as
% Let $\{\mybm{x}_t\}_{t\in [L]}$ represent an input sequence of row vectors $\mybm{x}_t \in \mathbb{R}^{1 \times D}$, where $[L] = \{1,\dots,L\}$, 
 % \todo{We seem to be mixing abbreviations for SSM and Selective SSM. Should we start talking about selective SSM from the very beginning?}
% $L$ is the sequence length, and $D$ is the feature size. 
% Selective SSM (S6)\footnote{Hereinafter, SSM refers to the S6 block in~\cite{gu2023mamba}.}~\cite{gu2023mamba} consists of an implicit hidden state $\mybm{h}_t \in \mathbb{R}^{N \times D}$ and four learnable continuous parameters $\mybm{A} \in \mathbb{R}^{N \times N}$, $\myDelta_t \in \mathbb{R}^{1 \times D}$, $\mybm{B}_t \in \mathbb{R}^{N \times 1}$, and $\mybm{C}_t \in \mathbb{R}^{1 \times N}$ with $t \in [L]$. Among these, only $\mybm{A}$ is independent of $\mybm{x}_t$. $\mybm{B}_t$, $\mybm{C}_t$, and $\myDelta_t$ are computed as
% \begin{align}
%     \mybm{B}_t &= \mathrm{Linear}_N(\mybm{x}_t) \\
%     \mybm{C}_t &= \mathrm{Linear}_N(\mybm{x}_t) \\
%     \myDelta_t &= \mathrm{Softplus}(s_{\myDelta}(\mybm{x}_t)),
% \end{align}
% where $s_\myDelta(\mybm{x}_t) = \mathrm{Broadcast}_D(\mathrm{Linear}_1(\mybm{x}_t))$, and $\mathrm{Linear}_d(\cdot)$ is a parameterized projection to a $d$-dimensional space.
As the objective was to make S6 work on discrete sequences of tokens (e.g. text)~\cite{gu2023mamba}, we have to discretize the continuous parameters $\mybm{A}$ and $\mybm{B}$ using discretization functions $f_A$ and $f_B$, that is $\overline{\mybm{A}} = f_A(\Delta, \mybm{A})$ and $\overline{\mybm{B}} = f_B(\Delta, \mybm{B})$, respectively.
% \begin{align}
%     \overline{\mybm{A}_t} &= f_A(\Delta_t, \mybm{A}) \\ %\exp(\myDelta_t \mybm{A}) \\
%     \overline{\mybm{B}_t} &= f_B(\Delta_t, \mybm{B}_t) %(\myDelta_t \mybm{A})^{-1}(\overline{\mybm{A}_t} - \mathbb{I}) \cdot \myDelta \mybm{B}_t,
% \end{align}
Finally, we can compute the hidden state $\mybm{h}_t$ and the output $\mybm{y}_t$ as follows
\begin{align}
    \mybm{h}_t &= \overline{\mybm{A}} \mybm{h}_{t-1} + \overline{\mybm{B}} \mybm{x}_t \\
    \mybm{y}_t &= \mybm{C} \mybm{h}_t.
\end{align}
As S6 is our focus, the term SSM hereinafter refers specifically to the S6 block in~\cite{gu2023mamba}.

\begin{figure}[t]
    \centering
    \croppdf{figures/dc_mamba_block}
    \includegraphics[width=\textwidth]{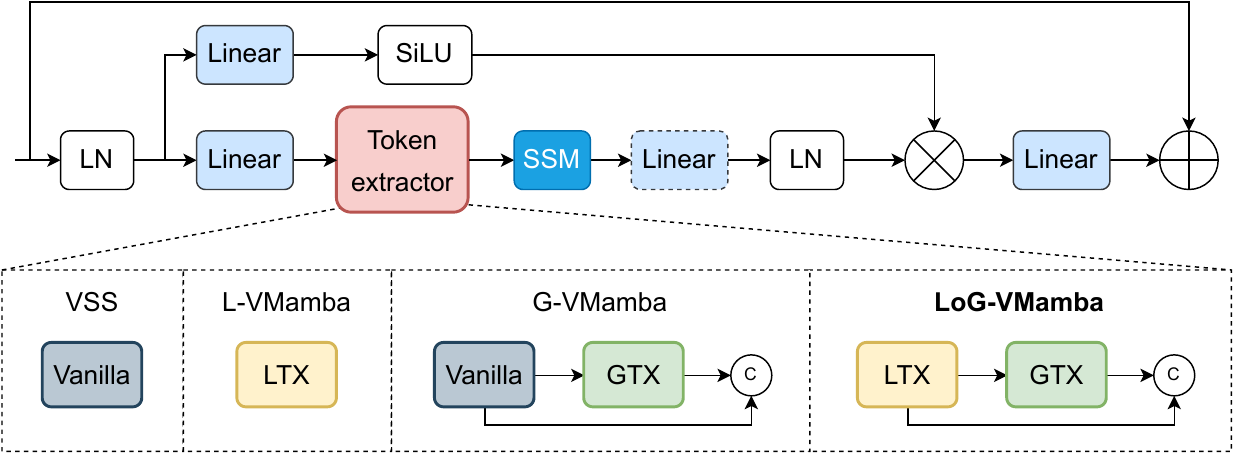}
    \caption{LoG-VMamba and its simpler versions compared to the vanilla VSS~\cite{liu2024vmamba}. LN and SSM mean layer normalization~\cite{ba2016layer} and the S6 block in~\cite{gu2023mamba}, respectively. Vanilla indicates the module consisting of a DWC layer and SiLU followed by a reshaping operator. The linear block after SSM is only needed in L-VMamba and LoG-VMamba. White blocks indicate modules without learnable parameters. $\bigoplus, \bigotimes$, and \textcopyright\ represent element-wise addition, multiplication, and concatenation, respectively. The SSM block in our settings performs only 1 horizontal scan.}
    \label{fig:mamba_blocks}
\end{figure}

\Paragraph{Visual State Space} 
(VSS)~\cite{liu2024vmamba} is an extension of the Mamba block~\cite{gu2023mamba} for visual data (see~\cref{fig:mamba_blocks}). Firstly, the input feature map $\mybm{x}$ is normalized and projected with an expansion factor $\alpha$ into two tensors going through two branches. In the first branch, one tensor is passed through a depthwise convolutional layer (DWC) and an activation function before getting processed by the SSM module in~\cite{gu2023mamba}. To enable SSM on 2D data, several scanning strategies are needed to convert a 2D array of tokens into a 1D sequence. Their required computational resources are proportional to the number of scanning directions $M$. After that, the output of the first branch is multiplied by that of the second branch, which has an activation function after the projection. The product is subsequently projected and added to the input $\mybm{x}$, resulting in the output of this block.
% \todo[inline]{x - in bold, right? Is index t still relevant? Fix / clarify.}
%---------------------------------------------------------------
\begin{figure}[tb]
  \centering
  \croppdf{figures/dc_global_local_extractors}
  \includegraphics[width=\textwidth]{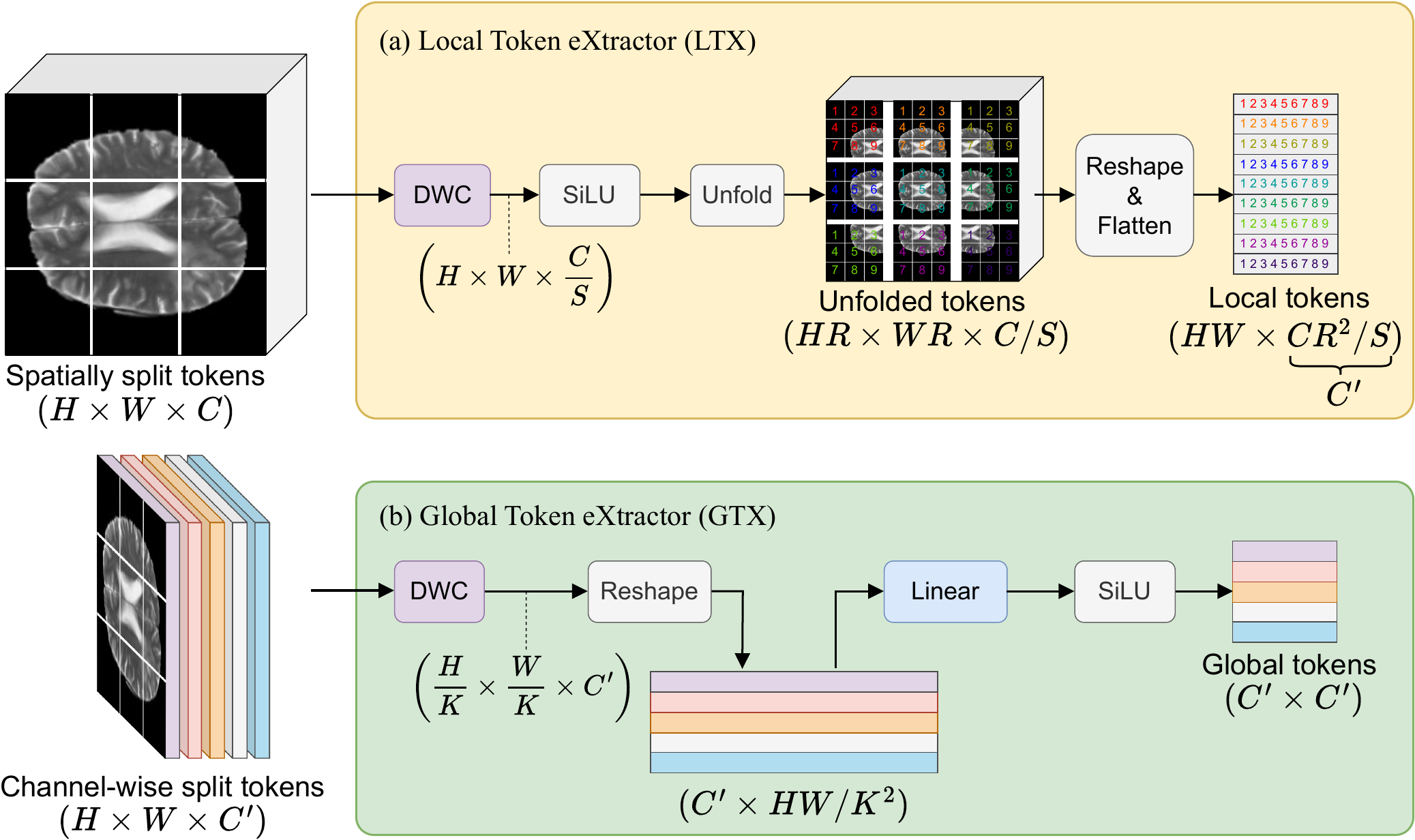}
  \caption{Local and global token extractors. DWC indicates a depthwise convolutional layer. $S$ and $K$ correspond to the depthwise and spatial compression in the DWCs of (a) and (b), respectively.}
  \label{fig:local_global_extractor}
\end{figure}

\subsection{Local-Global Token Extractors} 
\Paragraph{Local Token Extractor.}
As Mamba is a sequence modeling module, several scanning strategies~\cite{liu2024vmamba,zhu2024vision,huang2024localmamba,zhao2024rs} have been introduced to transform 2D arrays of tokens into 1D sequences.
However, their common pitfall, as illustrated in~\cref{fig:ss2d_h,fig:ss2d_v}, is that they fail to maintain the spatial proximity between neighboring tokens.
Meanwhile, the local dependencies are important to visual tasks~\cite{liu2021swin,huang2024localmamba}. 
% However, Mamba is a sequence modeling module, so we have to flatten the image patches into a sequence of visual tokens when applying the selective scan. Based on the scanning strategy that we use, neighboring tokens are not within the proximity of each other in the sequence.  
To improve the modeling of local features without resorting to an omnidirectional scan as in~\cite{zhao2024rs}, we introduce the LTX module, graphically illustrated in~\cref{fig:local_global_extractor}\textcolor{red}{a}. Specifically, we first utilize a DWC to squeeze the input channels $C$ by a factor $S$, which allows us to avoid the computational overhead of later operators. After passing the compressed tokens through SiLU, we use a fixed convolution kernel of size $R \times R$ to unfold the tokens~\cite{huang2023vision}. 
This operator allows us to replicate tokens and preserve the spatial relations of nearby tokens.
% Subsequently, we stack neighboring tokens within the local window along the channel axis of the query token. This pair of operators allows us to replicate tokens and preserve the spatial relations of nearby tokens. 
Finally, we reshape and flatten the spatial dimensions to form a 1D sequence of local tokens, ensuring that neighboring tokens within the local window are along the channel axis of the query token. This spatial flattening is performed in a row-by-row manner, conceptually equivalent to the horizontal scan in~\cref{fig:ss2d_h}. As a result, the number of output channels is $C' = \frac{C R^2}{S}$.

\Paragraph{Global Token Extractor.} In addition to the locality maintained by LTX, we propose the GTX module, demonstrated in~\cref{fig:local_global_extractor}\textcolor{red}{b}, to produce global (i.e.\ spatial-independent channel-wise) tokens. This module allows the SSM to access compressed versions of GRF at early time steps. Such an approach differs from
% scanning strategies
the selective scan
in prior VMs~\cite{liu2024vmamba,zhu2024vision,huang2024localmamba,zhao2024rs}, where only the token at the last time step has the context of all other tokens. Specifically, given an input feature map with dimensions $H\times W\times C'$, we spatially compress it using a dilated DWC with a stride of $K \times K$, and then flatten its spatial dimensions to produce global tokens of shape $C' \times \frac{HW}{K^2}$, with the channel dimension and spatial dimensions transposed. For computational efficiency, GTX merely compresses each group of $\gamma$ input channels throughout all spatial dimensions into global tokens. The role of these steps is to learn an approximation of the global context in each input channel, and thus the concern of losing fine-grained details associated with dilated DWC is insignificant. Subsequently, we utilize a linear layer to project these tokens to a $C'$-dimensional space and apply the activation function SiLU. Choosing the number of features as $C'$ enables us to concatenate the output of the GTX with the spatial tokens of the LTX of $C'$ channels in subsequent steps.

%---------------------------------------------------------------
% \todo{It feels like figures have redundant information (e.g. Fig 3 and Fig 4).}

%---------------------------------------------------------------
\subsection{Local-Global Vision Mamba}

We generalize the VSS block and incorporate the proposed LTX and GTX to introduce upgraded versions of VMamba, as graphically demonstrated in~\cref{fig:mamba_blocks}. As such, the \emph{vanilla} block consisting of a DWC layer and SiLU in VSS is treated as a token extraction module, leading us to develop the following Mamba-based models.

\Paragraph{Local Vision Mamba.} For Local Vision Mamba (L-VMamba), we employ the introduced LTX block as the token extractor. Compared to VSS, L-VMamba includes the unfolding operator to effectively guarantee the spatial proximity of neighboring tokens in 2D or 3D arrays. We set the window size $R = 3$, which is a common kernel size of convolutional layers. Since the number of channels is changed to $C'$ in the LTX, a linear layer is inserted after the SSM to project the tokens to the original $C$-dimensional space. 
% \todo{You explicitly say what is vanilla block only in the figure caption, but not in the text. It is mentioned very briefly in the beginning of the next paragraph, but it is not salient enough IMO. Also, I generally do not like the word ``vanilla'' in papers ==> Explained vanilla block in the overview paragraph. I used the term vanilla because we specifically use the vanilla version of VSS.}
% However, the tradeoff is the smaller number of features due to the preceding squeezing DWC. Our empirical evidence in~\cref{tab:ablation_components} shows that despite such a tradeoff, this simple approach results in significant improvement compared to VSS.

\Paragraph{Global Vision Mamba.} In Global Vision Mamba (G-VMamba), we combine the vanilla block, comprising a DWC layer and SiLU followed by a flattening operator, from VSS with the GTX block. After being processed by the vanilla block, the feature maps are passed through the GTX module to produce tokens with a GRF. The tokens produced by both modules are eventually combined and forwarded to the next SSM. 
% \todo{What is the meaning of this sentence? Do we need it? Feels weird. Which sentence?}
As the number of channels in the outputs is unchanged while the sequence length increases, the linear layer after the SSM is unnecessary.

% which are subsequently concatenated with the output of the vanilla block. 
% Later in this section, we will detail the approaches for concatenating the two types of tokens.

\Paragraph{Local-Global Vision Mamba.} 
\label{sc:log_vmamba}
Ultimately, we couple the proposed modules, LTX and GTX, to create the LoG-VMamba module. This combination leverages the local dependencies of LTX and the GRF of GTX, thus harnessing the strengths of both. Precisely, given an input feature map $\mybm{x} \in \mathbb{R}^{H \times W \times C}$, the token extractor of LoG-VMamba is formulated as% follows
\begin{align}
    \mybm{x}^L &= \mathrm{LTX}(\mybm{x}), \quad \quad \mybm{x}^G = \mathrm{GTX}(\mybm{x}^L) \\
    \mybm{x}^{LG} &= \mathrm{Concat}(\mybm{x}^G,\mybm{x}^L),
\end{align}
% where $\cup$ stands for concatenation of columns of two matrices, $\mathrm{LTX}: \mathbb{R}^{H \times W \times C} \rightarrow \mathbb{R}^{HW \times C'}$, $\mathrm{GTX}:\mathbb{R}^{H \times W \times C'} \rightarrow \mathbb{R}^{C' \times C'}$, 
where $\mybm{x}^L \in \mathbb{R}^{HW \times C'}$, $\mybm{x}^G \in \mathbb{R}^{C' \times C'}$, $\mybm{x}^{LG} \in \mathbb{R}^{(HW + C')\times C'}$, and $C'=\frac{CR^2}{S}$. Similar to L-VMamba, a linear layer after the SSM module is needed to map the output $\mybm{x}^{LG}$ to a $C$-dimensional space.

%As with L-VMamba, following the SSM, a linear layer is needed to map the output $\mybm{x}^{LG}$ to a $C$-dimensional space.
Due to the sequential and input-dependent nature of Mamba, the concatenation between $\mybm{x}^L$ with $\mybm{x}^G$ is not trivial in how to harness both types of contexts. Therefore, we evaluate the following approaches:
\begin{itemize}
    \item Head: Concatenating the global tokens at the beginning of the sequence.
    \item Middle: Placing the global tokens in the middle of the sequence.
    \item Split: Dividing the global tokens into two halves and appending them to both ends of the sequence.
    \item Interleaved: Inserting each global token between the local tokens at fixed intervals over the sequence, with excessive global tokens at the beginning.
\end{itemize}

% Thanks to the enriched tokens produced by our method, multiple scans in different directions in the SSM module are unnecessary. Instead, we merely perform a simple horizontal scan (see~\cref{fig:ss2d_h}), significantly improving computational efficiency.
% \todo{Mention why it would make sense to use different concatenation strategy. One may wonder. ATM it does not feel like it has any value. At least based on the ablation. I guess you started this because of the sequential nature of mamba, as it may forget the global information of an image is e.g. very big. => It affects how often the sequence can access GRF. Ultimately it cannot be an arbitrary strategy, so we compare a few and choose 'Interleaved' as the default.}

\begin{table}[t]
  \centering
  \caption{Configurations of the experimental setup for different datasets}
  \label{tab:configs}
  \begin{tabular}{|l|c|c|c|c|c|c|c|}
  \hline
  \multicolumn{1}{|c|}{Dataset} & Image size & Batch size & Weight decay & Stages & Pooling & $K$ & $S$ \\
  \hline
  Endoscopy & $640\times384$ & $8$ & $5 \cdot 10^{-2}$ & $4$ & $5\times5$ & $2\times2$ & $8$ \\
  Cell & $512\times512$ & $8$ & $5 \cdot 10^{-2}$ & $4$ & $5\times5$ & $2\times2$ & $8$ \\
  BraTS & $128\times128\times128$ & $2$ & $10^{-4}$ & $5$ & $5\times5\times5$ & $2\times2\times2$ & $16$ \\
  ACDC & $16\times256\times224$ & $2$ & $10^{-4}$ & $6$ & $3\times5\times5$ & $2\times4\times4$ & $16$ \\
  \hline
  \end{tabular}
\end{table}

%---------------------------------------------------------------
\subsection{Medical Image Segmentation Models}

% \todo{Rewrite: explain first how you implement the design, and then refer to supplementary. }
With the proposed Mamba-based modules, we present our segmentation models for both 2D and 3D medical imaging data. For 2D segmentation, we build our model on top of Swin-UMamba$^\dagger$~\cite{liu2024swin}. Different from Swin-UMamba, the decoder of Swin-UMamba$^\dagger$ comprises VSS blocks instead of CNNs. As empirically shown in~\cite{liu2024swin}, the pre-trained weights in its encoder are of significant importance,
we thus retain the original encoder. In the decoder, we replace the original VSS blocks with our LoG-VMamba blocks. For 3D segmentation models, we modify a version of U-Mamba-Enc~\cite{ma2024u}. In the encoder, we replace the Mamba blocks in the original network with the LoG-VMamba blocks. The decoder remains unchanged as it does not incorporate any Mamba modules. Following the methods in~\cite{liu2024swin,ma2024u}, we tailor our models accordingly for each dataset. The details of the 2D and 3D segmentation models are graphically depicted in~\cref{fig:dc_segmentation}.

% To leverage the VMamba~\cite{liu2024vmamba} backbone in Swin-UMamba$^\dagger$, which is pretrained on the ImageNet dataset~\cite{russakovsky2015imagenet}, 
% Following~\cite{liu2024swin,ma2024u}, we change the specific details of our model on each dataset to match their configurations.

%===========================================================
\section{Experiments}

%---------------------------------------------------------------
\subsection{2D Datasets}
 We followed prior works~\cite{ma2024u,liu2024swin} and employed these two 2D segmentation datasets for our experiments: Endoscopy from the MICCAI 2017 EndoVis Challenge~\cite{allan20192017} and Cell from the NeurIPS 2022 Cell Segmentation
Challenge~\cite{ma2024multimodality}. 

\textbf{Endoscopy} contained $1800$ training and $1200$ test images. We further split the training set into $1440$ and $360$ samples for training and validation, respectively. Its objective was to segment seven instruments: prograsp forceps, needle driver, monopolar curved scissors, bipolar forceps, cadiere forceps, vessel sealer, and drop-in ultrasound probe. 

\textbf{Cell} consisted of $1000$ and $101$ samples for training and evaluation, respectively. In addition, we divided the former into two portions of size $800$ and $200$ for training and validation, respectively. We performed both semantic and instance segmentation of cells on this dataset. 

\begin{figure}[t]
    \centering
    \croppdf{figures/dc_model_outputs}
    \includegraphics[width=\textwidth]{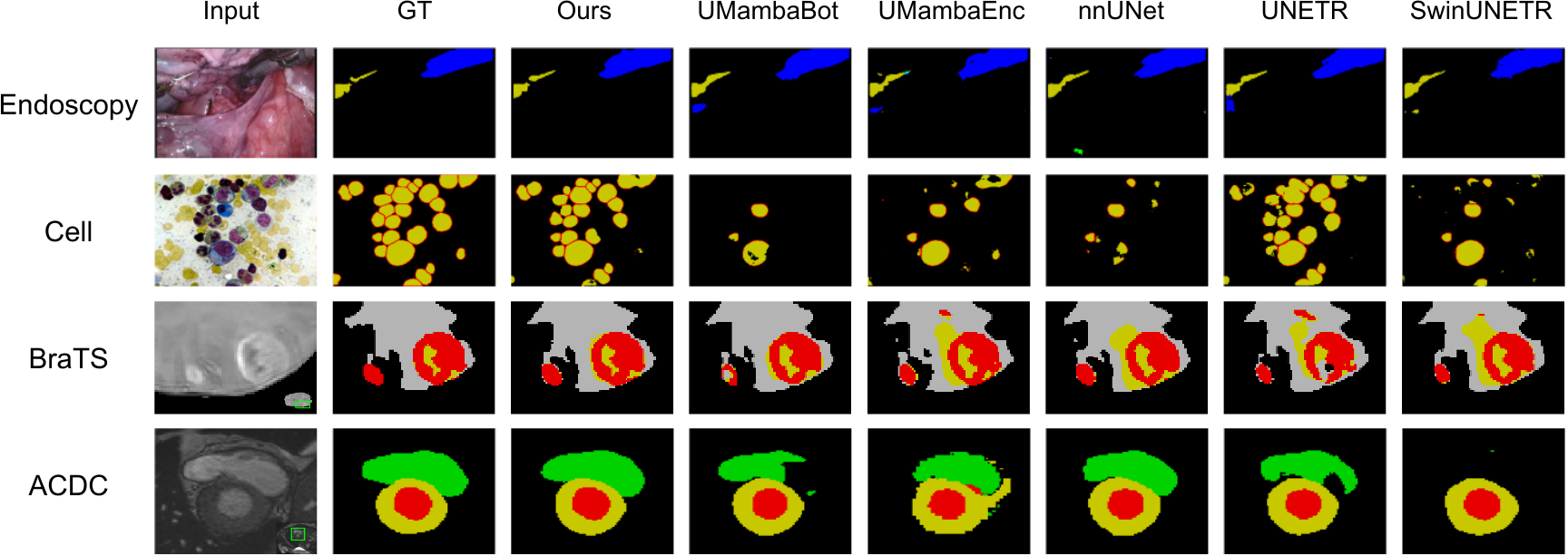}
    \caption{Qualitative comparisons between our method and the baselines}
    \label{fig:viz_results}
\end{figure}

% \begin{figure}[t]
%     \centering
%     \croppdf{figures/radar_chart}    
%     \subfloat[Radar chart]{\includegraphics[width=0.47\textwidth]{figures/radar_chart-crop.pdf}}
%     \hfill
%     \subfloat[Radar chart]{\includegraphics[width=0.47\textwidth]{figures/radar_chart-crop.pdf}}
%     \caption{Performance.}
%     \label{fig:performance_overall}
% \end{figure}

\subsection{3D Datasets} We conducted experiments on two 3D segmentation datasets: BraTS 2020~\cite{menze2014multimodal,bakas2017advancing,bakas2018identifying} and ACDC~\cite{bernard2018deep}. 

\textbf{BraTS} consisted of multi-modal magnetic resonance (MR) images collected from $369$ subjects. We split this dataset into training, validation, and test sets with $236$, $59$, and $74$ MR images, respectively. The common volume size was $240 \times 240 \times 155$. Following the literature, we focused on three objects of interest: enhancing tumor (ET), tumor core (TC), and whole tumor (WT). 

\textbf{ACDC} was composed of a training and test split, whose size was $200$ and $100$ samples, respectively. The former was further divided into $160$ training and $40$ validation images. The segmentation targets were right ventricle (RV), left ventricle (LV), and myocardium (MYO).

\subsection{Implementation Details} 
\textbf{Model Training and Hyperparameters.} Our models were trained using Nvidia V100 GPUs. We implemented our methods using Pytorch~\cite{paszke2019pytorch}. On each dataset, we followed a standard data preprocessing pipeline for all models. While training on 2D inputs, we extracted patches of a standard size, followed by augmentations by flipping, elastic deformation, color jittering, and noise addition. During the optimization on 3D data, the MR images were randomly cropped to a fixed volume size, and then augmented by flipping, intensity scaling, and intensity shifting. We applied a sliding window in the validation and test stages.

Our 2D and 3D segmentation models used $\alpha$'s of $2$ and $1$, respectively. We merely performed $1$ horizontal scan ($M=1$) for our methods. In our GTX blocks, except for the highest level that used $2$ channels per token ($\gamma=2$), other global tokens corresponded to $1$ channel ($\gamma=1$). Among different concatenation strategies, the ``Interleaved'' setting was empirically chosen. To train our proposed models, we used the Adam optimizer~\cite{kingma2014adam} with an initial learning rate of $10^{-4}$. We utilized a threshold of $0.5$ to binarize predictions. The more specific configurations are presented in~\cref{tab:configs}. Our employed loss function was the sum of Dice and Cross-Entropy losses. All experiments were repeated using $5$ different random seeds and folds, and we reported the means and standard errors across these $5$ runs.

\begin{figure}[t]
    \centering
    \croppdf{figures/flops_perf_endoscopy}
    \croppdf{figures/flops_perf_cell}
    \hspace*{\fill}
    \subfloat[Endoscopy dataset\label{fig:2d_comparisons_endoscopy}]{\includegraphics[width=0.45\textwidth]{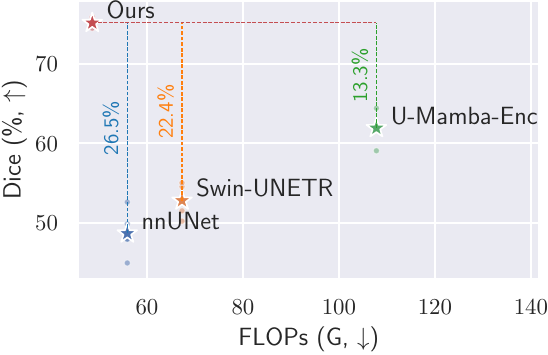}}
    \hfill
    \subfloat[Cell dataset\label{fig:2d_comparisons_cell}]{\includegraphics[width=0.45\textwidth]{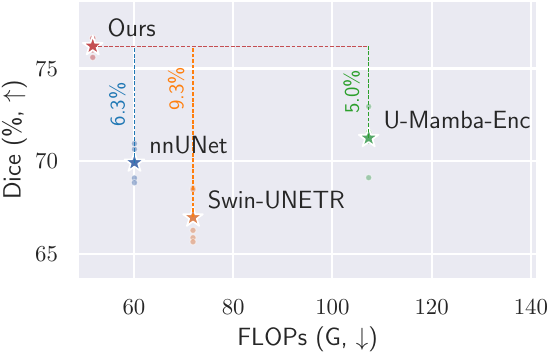}}
    \hspace*{\fill}
    \caption{Computational efficiency and performance comparisons on the 2D datasets. Stars indicate the means while blurry dots represent the individual results.}
    \label{fig:2d_comparisons}
\end{figure}

\begin{figure}[t]
    \centering
    \croppdf{figures/ACDC_radar}
    \croppdf{figures/BraTS_radar}
    \hspace*{\fill}
    \subfloat[BraTS dataset\label{fig:3d_comparison_brats}]{\includegraphics[height=0.35\textwidth]{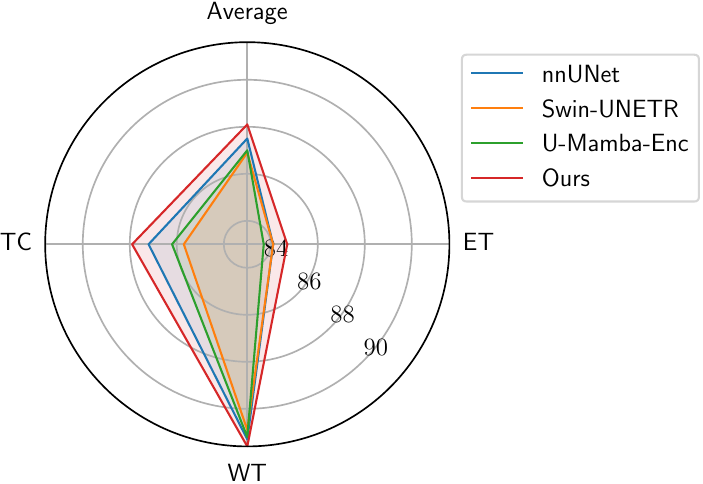}}
    \hfill
    \subfloat[ACDC dataset\label{fig:3d_comparison_acdc}]{\includegraphics[height=0.35\textwidth]{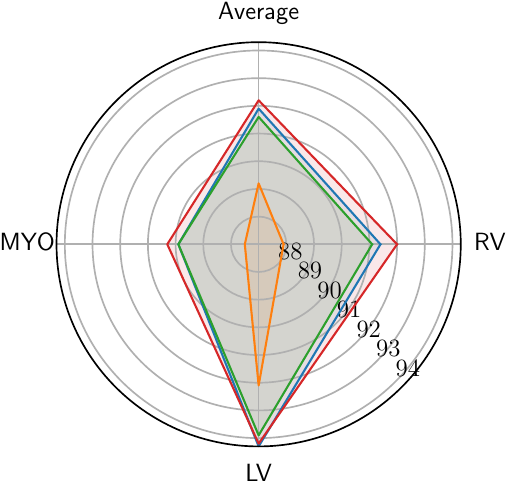}}
    \hspace*{\fill}
    \caption{Fine-grained performance comparisons on the 3D datasets (Dice, \% $\uparrow$)}
    \label{fig:3d_comparisons}
\end{figure}

\Paragraph{Metrics.} We evaluated 2D models with Dice score and Intersection-over-Union (IoU). 
% On the Cell dataset, we reported both instance and semantic segmentation results. 
Regarding 3D models, we computed the Dice score for each class. In addition, we employed one surface-distance-based performance metric. We used Normalized Surface Dice (NSD) for 2D models following~\cite{ma2024u,liu2024swin} and 95\% Hausdorff distance (HD95) following~\cite{xing2022nestedformer,she2023eoformer,dang2024singr}.

% U-Mamba-Bot Enc: 1
% Swin-Umamba: 4
% SegMamba: 3
%---------------------------------------------------------------
\subsection{Comparison with State-of-The-Art Methods}
% \todo{Add a radar plot per metric. Visualize at most three models in the plot (two is the best number here, I think). Your axes on the plot are datasets. }

We compared our method to a diverse array of references on both 2D and 3D medical imaging datasets. These references included convolutional, Transformer-based, and Mamba-based models, whose representatives were nnUNet~\cite{isensee2024nnu}, Swin-UNETR~\cite{hatamizadeh2021swin}, and U-Mamba-Enc~\cite{ma2024u}, respectively. In general, our proposed method consistently acquired improvements over the baselines across $4$ different datasets without compromising efficiency. The quantitative comparisons are presented in~\cref{fig:2d_comparisons,fig:3d_comparisons},~\cref{tab:endovis_results,tab:cell_results,tab:brats_results,tab:acdc_results}, and~\cref{tab:brats_class_results,tab:acdc_class_results}. The qualitative results are demonstrated in~\cref{fig:viz_results}.

\begin{table}[tb]
\centering
\renewcommand{\arraystretch}{1.2}
\caption{Comparisons on the Endoscopy test set. The best results are highlighted in bold.}
\label{tab:endovis_results}
\begin{tabular}{|l|r|r|c|c|c|c|}
\hline 
\multicolumn{1}{|c|}{Method} & \multicolumn{1}{c}{Size} & \multicolumn{1}{|c|}{FLOPs} & Dice (\%) $\uparrow$ & IoU (\%) $\uparrow$ & NSD (\%) $\uparrow$ \\
\hline  % \midrule   
UNet~\cite{ronneberger2015u} & $32.5$M & $40.8$G & 32.61$_{\pm2.67}$ & 29.00$_{\pm2.67}$ & 33.69$_{\pm2.69}$ \\
nnUNet~\cite{isensee2024nnu} & $47.6$M & $55.9$G & 48.64$_{\pm1.13}$ & 45.21$_{\pm1.09}$ & 49.88$_{\pm1.15}$ \\
UNETR~\cite{hatamizadeh2022unetr} & $88.3$M & $111.7$G & 41.87$_{\pm0.49}$ & 38.45$_{\pm0.48}$ & 43.27$_{\pm0.50}$ \\
Swin-UNETR~\cite{hatamizadeh2021swin} & $25.1$M & $67.3$G & 52.78$_{\pm0.81}$ & 49.55$_{\pm0.78}$ & 54.31$_{\pm0.83}$ \\
\hline
U-Mamba-Bot~\cite{ma2024u} & $64.0$M & $99.3$G & 62.20$_{\pm1.85}$ & 58.11$_{\pm1.93}$ & 63.78$_{\pm1.85}$ \\
U-Mamba-Enc~\cite{ma2024u} & $66.0$M & $107.8$G & 61.91$_{\pm0.77}$ & 57.85$_{\pm0.76}$ & 63.43$_{\pm0.77}$ \\
Swin-UMamba~\cite{liu2024swin} & $59.9$M & $163.7$G & 65.15$_{\pm0.70}$ & 61.65$_{\pm0.72}$ & 66.64$_{\pm0.71}$ \\
Swin-UMamba$^\dagger$~\cite{liu2024swin} & $27.5$M & $45.4$G & 71.23$_{\pm1.00}$ & 67.81$_{\pm0.99}$ & 72.77$_{\pm1.02}$ \\
% \hline
\rowcolor{LightCyan}
Ours & $30.3$M & $48.6$G & \subbest{75.17$_{\pm0.24}$} & \subbest{71.68$_{\pm0.23}$} & \subbest{76.83$_{\pm0.25}$} \\
% UNet~\cite{ronneberger2015u} & $32.5$M & $40.8$G & $40.61 \pm 0.53$ & $37.00 \pm 0.50$ & $41.78 \pm 0.52$ \\
% nnUNet~\cite{isensee2024nnu} & $47.6$M & $55.9$G & $48.64 \pm 0.50$ & $45.21 \pm 0.49$ & $49.88 \pm 0.51$ \\
% UNETR~\cite{hatamizadeh2022unetr} & $88.3$M & $111.7$G & $43.45 \pm 0.24$ & $40.06 \pm 0.24$ & $44.85 \pm 0.25$ \\
% Swin-UNETR~\cite{hatamizadeh2021swin} & $25.1$M & $67.3$G & $55.01 \pm 0.90$ & $51.69 \pm 0.90$ & $56.59 \pm 0.90$ \\
% \hline
% U-Mamba-Bot~\cite{ma2024u} & $64.0$M & $99.3$G & $64.96 \pm 0.69$ & $60.97 \pm 0.62$ & $66.56 \pm 0.70$ \\
% U-Mamba-Enc~\cite{ma2024u} & $66.0$M & $107.8$G & $64.42 \pm 0.25$ & $60.33 \pm 0.25$ & $65.96 \pm 0.25$ \\
% Swin-UMamba~\cite{liu2024swin} & $59.9$M & $163.7$G & $65.15 \pm 0.31$ & $61.65 \pm 0.32$ & $66.64 \pm 0.32$ \\
% Swin-UMamba$^\dagger$~\cite{liu2024swin} & $27.5$M & $45.4$G & $71.23 \pm 0.45$ & $67.81 \pm 0.44$ & $72.77 \pm 0.45$ \\
% \hline
% Ours & $30.3$M & $44.3$G & $75.17 \pm 0.11$ & $71.68 \pm 0.10$ & $76.83 \pm 0.11$ \\
\hline
\end{tabular}
\end{table}

\Paragraph{Endoscopy.} In~\cref{fig:2d_comparisons_endoscopy}, we compared our method to three representative baselines: nnUNet (CNN-based)~\cite{isensee2024nnu}, Swin-UNETR (Transformer-based)~\cite{hatamizadeh2021swin}, and U-Mamba-Enc (SSM-based)~\cite{ma2024u}. The results demonstrate that our method overcame the typical trade-off between computational cost and performance among the baselines.
% \todo{'disrupted' sounds weird.} 
While achieving the lowest FLOPs, it significantly outperformed the three baselines with differences of $26.5\%$, $22.4\%$, and $13.3\%$ in Dice, respectively. 
% Am, the best-performing baseline was Swin-UMamba$^\dagger$, served as the foundation for our segmentation model.
% with $71.23\%$ Dice, $67.81\%$ IoU, and $72.77\%$ NSD. 
% This is also the model that we used as the reference for our segmentation model.
% , on which we replaced the original Mamba blocks with our proposed modules. 
% By simply using the LoG-VMamba blocks, we observed a significant increase in performance metrics: $3.94\%$ in Dice, $3.87\%$ in IoU, and $4.06\%$ in NSD.
Compared to the best-performing baseline, Swin-UMamba$^\dagger$~\cite{liu2024swin}, our method led to significant improvements of $3.94\%$ in Dice, $3.87\%$ in IoU, and $4.06\%$ in NSD. More quantitative results are presented in~\cref{tab:endovis_results}.
% , leading to $75.17\%$ Dice and $76.83\%$ NSD. 
% As a result, our segmentation model achieved the highest performance on the Endoscopy dataset, in terms of Dice, IoU, and NSD. 
% In addition, our method did not require a huge number of model parameters and FLOPs to deliver competitive results. 
% Compared to the original Swin-UMamba$^\dagger$, our model used only $2.8$M more parameters and $3.2$G more FLOPs, which was the third lowest in terms of model size and FLOPs.

\begin{table}[tb]
\centering
\renewcommand{\arraystretch}{1.2}
\caption{Comparisons on the Cell test set. Metric\textsuperscript{i} denotes an instance segmentation metric while Metric\textsuperscript{s} denotes a semantic segmentation metric. The best results are highlighted in bold.}
\label{tab:cell_results}
\begin{tabular}{|l|r|r|c|c|c|c|c|}
\hline 
\multicolumn{1}{|c|}{Method} & \multicolumn{1}{c}{Size} & \multicolumn{1}{|c|}{FLOPs} & Dice\textsuperscript{i} (\%) $\uparrow$ & IoU\textsuperscript{i} (\%) $\uparrow$ & Dice\textsuperscript{s} (\%) $\uparrow$ & NSD\textsuperscript{s} (\%) $\uparrow$ \\
\hline  % \midrule      
UNet~\cite{ronneberger2015u} & $32.5$M & $43.1$G & 41.23$_{\pm1.35}$ & 30.99$_{\pm1.11}$ & 53.00$_{\pm1.50}$ & 61.87$_{\pm1.70}$ \\
nnUNet~\cite{isensee2024nnu} & $65.2$M & $60.1$G & 53.44$_{\pm0.56}$ & 41.41$_{\pm0.54}$ & 69.92$_{\pm0.37}$ & 79.03$_{\pm0.41}$ \\
UNETR~\cite{hatamizadeh2022unetr} & $88.3$M & $120.3$G & 44.42$_{\pm0.79}$ & 32.52$_{\pm0.67}$ & 71.28$_{\pm0.86}$ & 80.03$_{\pm0.65}$ \\
Swin-UNETR~\cite{hatamizadeh2021swin} & $25.1$M & $71.9$G & 42.90$_{\pm0.26}$ & 31.91$_{\pm0.25}$ & 66.95$_{\pm0.57}$ & 76.48$_{\pm0.54}$ \\
\hline
U-Mamba-Bot~\cite{ma2024u} & $86.2$M & $101.8$G & 57.86$_{\pm0.50}$ & 45.16$_{\pm0.55}$ & 70.86$_{\pm0.48}$ & 80.52$_{\pm0.43}$ \\
U-Mamba-Enc~\cite{ma2024u} & $86.4$M & $107.3$G & 59.50$_{\pm0.47}$ & 46.99$_{\pm0.49}$ & 71.25$_{\pm0.55}$ & 80.54$_{\pm0.64}$ \\
Swin-UMamba~\cite{liu2024swin} & $59.9$M & $174.5$G & 52.92$_{\pm1.02}$ & 40.02$_{\pm0.99}$ & 72.52$_{\pm0.92}$ & 82.80$_{\pm0.92}$ \\
Swin-UMamba$^\dagger$~\cite{liu2024swin} & $27.5$M & $48.3$G & 57.89$_{\pm1.17}$ & 45.21$_{\pm1.11}$ & 73.50$_{\pm0.86}$ & 83.31$_{\pm0.66}$ \\
% \hline
\rowcolor{LightCyan}
Ours & $30.3$M & $51.7$G & \subbest{60.74$_{\pm0.27}$} & \subbest{47.88$_{\pm0.32}$} & \subbest{76.21$_{\pm0.10}$} & \subbest{86.44$_{\pm0.09}$} \\
% UNet~\cite{ronneberger2015u} & $32.5$M & $43.1$G & $44.54 \pm 0.14$ & $33.53 \pm 0.18$ & $57.23 \pm 1.00$ & $67.36 \pm 1.00$ \\
% nnUNet~\cite{isensee2024nnu} & $65.2$M & $60.1$G & $53.44 \pm 0.32$ & $41.41 \pm 0.30$ & $69.92 \pm 0.21$ & $79.03 \pm 0.23$ \\
% UNETR~\cite{hatamizadeh2022unetr} & $88.3$M & $120.3$G & $46.42 \pm 0.46$ & $34.26 \pm 0.38$ & $73.45 \pm 0.36$ & $81.77 \pm 0.36$ \\
% Swin-UNETR~\cite{hatamizadeh2021swin} & $25.1$M & $71.9$G & $43.86 \pm 0.18$ & $32.57 \pm 0.19$ & $68.46 \pm 0.34$ & $78.10 \pm 0.38$ \\
% \hline
% U-Mamba-Bot~\cite{ma2024u} & $86.2$M & $101.8$G & $57.86 \pm 0.28$ & $45.16 \pm 0.31$ & $70.86 \pm 0.27$ & $80.52 \pm 0.24$ \\
% U-Mamba-Enc~\cite{ma2024u} & $86.4$M & $107.3$G & $59.50 \pm 0.26$ & $46.99 \pm 0.27$ & $71.25 \pm 0.31$ & $80.54 \pm 0.36$ \\
% Swin-UMamba~\cite{liu2024swin} & $59.9$M & $174.5$G & $52.92 \pm 0.57$ & $40.02 \pm 0.56$ & $72.52 \pm 0.52$ & $82.80 \pm 0.51$ \\
% Swin-UMamba~\cite{liu2024swin} & $27.5$M & $48.3$G & $57.89 \pm 0.65$ & $45.21 \pm 0.48$ & $73.50 \pm 0.62$ & $83.31 \pm 0.37$ \\
% \hline
% Ours & $30.3$M & $47.1$G & $59.96 \pm 0.43$ & $47.54 \pm 0.40$ & $75.65 \pm 0.31$ & $85.58 \pm 0.32$ \\
\hline
\end{tabular}
\end{table}

\Paragraph{Cell.} For the Cell dataset, we evaluated metrics for both instance and semantic segmentation in~\cref{fig:2d_comparisons_cell,tab:cell_results}. Each model predicted regions of cells, as well as their boundaries. These boundaries were used to separate the cell foreground into multiple cell instances.
% The Mamba-based architectures achieved competitive performances compared to traditional convolutional and Transformer-based baselines.
Compared to nnUNet~\cite{isensee2024nnu}, Swin-UNETR~\cite{hatamizadeh2021swin}, and U-Mamba-Enc~\cite{ma2024u} on the semantic segmentation task, our method was the most computationally efficient while surpassing these baselines with differences of $6.3\%$, $9.3\%$, and $5.0\%$ in Dice, respectively (see~\cref{fig:2d_comparisons_cell}). For both instance and semantic segmentation, our model consistently achieved the highest performances across all the metrics. Compared to Swin-UMamba$^\dagger$~\cite{liu2024swin}, on which our model was based, using LoG-VMamba resulted in substantial gains of $2.71\%$ in Dice and $3.13\%$ in NSD for semantic segmentation, as well as $2.85\%$ in Dice and $2.67\%$ in IoU for instance segmentation.
% Compared to Our model also showed an improvement of $1.24\%$ in Dice and $0.89\%$ in IoU over U-Mamba-Enc for instance segmentation while using less parameters and FLOPs.
% U-Mamba-Enc reached $59.50\%$ in Dice and $46.99\%$ in IoU on instance segmentation, which was the highest among all baselines. 
% Regarding semantic segmentation results, Swin-UMamba$^\dagger$ was the baseline that showed the best performance with $73.5\%$ in Dice and $83.31\%$ in NSD. 
% Our model once again achieved a better performance than all the baselines for both instance and semantic segmentation. We built our model on top of Swin-UMamba$^\dagger$, and it resulted in an increase of $2.71\%$ in Dice and $3.13$ in NSD for semantic segmentation. Our model also showed an improvement of $1.24\%$ in Dice and $0.89\%$ in IoU over U-Mamba-Enc for instance segmentation while using less parameters and FLOPs, as illustrated in~\cref{fig:2d_comparisons}.

\begin{table}[t]
\centering    
\renewcommand{\arraystretch}{1.2}
\caption{Comparisons on the BraTS test set. The best results are highlighted in bold.}
\label{tab:brats_results}
\begin{tabular}{|l|r|r|c|c|c|c|c|c|}
\hline 
\multicolumn{1}{|c|}{Method} & \multicolumn{1}{c|}{\multirow{1}{*}{Size}} & \multicolumn{1}{c|}{\multirow{1}{*}{FLOPs}} & Dice score (\%) $\uparrow$ & HD95 ($mm$) $\downarrow$ \\
\hline
% TransBTS~\cite{wang2021transbts} & $33.0$M & $327.1$G & 84.77$_{\pm0.21}$ & 5.68$_{\pm0.18}$ \\
% SegResnet~\cite{myronenko20193d} & $1.2$M & $38.6$G & 85.78$_{\pm0.18}$ & 4.85$_{\pm0.21}$ \\
UNet3D~\cite{kerfoot2019left} & $5.7$M & $306.1$G & 86.52$_{\pm0.12}$ & 5.42$_{\pm0.37}$ \\
nnUNet~\cite{isensee2024nnu} & $192.2$M & $727.5$G & 
87.54$_{\pm0.26}$ & 4.83$_{\pm0.44}$ \\
UNETR~\cite{hatamizadeh2022unetr} & $102.4$M & $184.9$G & 84.78$_{\pm0.24}$ & 5.90$_{\pm0.22}$ \\
Swin-UNETR~\cite{hatamizadeh2021swin} & $62.2$M & $777.3$G & 86.90$_{\pm0.20}$ & 5.36$_{\pm0.20}$ \\
NestedFormer~\cite{xing2022nestedformer} & $10.6$M & $207.3$G & 86.71$_{\pm0.05}$ & 6.12$_{\pm0.39}$ \\
EoFormer~\cite{she2023eoformer}& $6.4$M & $81.7$G & 86.18$_{\pm0.18}$ & 5.34$_{\pm0.24}$ \\
\hline
U-Mamba-Bot~\cite{ma2024u} & $30.1$M & $393.9$G & 87.54$_{\pm0.18}$ & 4.30$_{\pm0.16}$ \\
U-Mamba-Enc~\cite{ma2024u} & $32.0$M & $401.2$G & 87.01$_{\pm0.10}$ & 4.38$_{\pm0.09}$ \\
SegMamba~\cite{xing2024segmamba} & $66.9$M & $1477.3$G & 87.62$_{\pm0.16}$ & 4.73$_{\pm0.22}$ \\
% \hline
\rowcolor{LightCyan}
Ours & $31.5$M & $401.6$G & \subbest{88.06$_{\pm0.08}$} & \subbest{3.97$_{\pm0.04}$} \\
% TransBTS~\cite{wang2021transbts} & $33.0$M & $327.1$G & $84.8 \pm 0.2$ & $5.7 \pm 0.2$ \\
% SegResnet~\cite{myronenko20193d} & $1.2$M & $38.6$G & $85.8 \pm 0.2$ & $4.9 \pm 0.2$ \\
% UNETR~\cite{hatamizadeh2022unetr} & $102.4$M & $184.9$G & $84.8 \pm 0.2$ & $5.9 \pm 0.2$ \\
% EoFormer~\cite{she2023eoformer}& $6.4$M & $81.7$G & $86.2 \pm 0.2$ & $5.3 \pm 0.2$ \\
% UNet3D~\cite{kerfoot2019left} & $5.7$M & $306.1$G & $86.5 \pm 0.1$ & $5.4 \pm 0.4$ \\  
% NestedFormer~\cite{xing2022nestedformer} & $10.6$M & $207.3$G & $86.7 \pm 0.1$ & $6.1 \pm 0.4$ \\
% Swin-UNETR~\cite{hatamizadeh2021swin} & $15.7$M & $207.5$G & $86.9 \pm 0.2$ & $5.4 \pm 0.2$ \\
% U-Mamba-Bot~\cite{ma2024u} & $30.1$M & $393.9$G & $87.58 \pm 0.3$ & $4.16 \pm 0.18$ \\
% U-Mamba-Enc~\cite{ma2024u} & $32.0$M & $401.2$G & $86.90 \pm 0.05$ & $4.48 \pm 0.11$ \\
% \hline
% Ours & $31.5$M & $400.2$G & $88.03 \pm0.26$ & $3.97 \pm 0.04$ \\
\hline
\end{tabular}
\end{table}

\Paragraph{BraTS 2020.} Apart from the gain in performances on 2D datasets, our method also showed competitive results on 3D datasets such as BraTS.
% , shown in~\cref{tab:brats_results}. 
% We also present fine-grained comparisons in~\cref{tab:brats_class_results,fig:3d_comparisons}. 
In~\cref{fig:3d_comparison_brats}, our method substantially performed better than the three baselines -- nnUNet~\cite{isensee2024nnu}, Swin-UNETR~\cite{hatamizadeh2021swin}, and U-Mamba-Enc~\cite{ma2024u} -- on all individual classes. When we directly compared our method to its base model, U-Mamba-Enc, we observed substantial improvements of $1.05\%$ in Dice and $0.41$mm in HD95. We present more quantitative results in~\cref{tab:brats_results,tab:brats_class_results}. As we used an expansion factor $\alpha = 1$ in every Mamba block, we spent $0.5$M fewer parameters than U-Mamba-Enc while using a similar number of FLOPs. Ultimately, compared to SegMamba~\cite{xing2024segmamba}, the most competitive baseline, our method obtained better results with differences of $0.44\%$ in Dice and $0.76$mm in HD95. 
% Among the baselines, SegMamba delivered a Dice score of $87.62\%$ and U-Mamba-Bot obtained a HD95 of $4.30$mm. 
% We based our segmentation model on U-Mamba-Enc and replaced every Mamba block in the encoder with our proposed module. These changes led to $88.06\%$ Dice and $3.97$mm HD95, which is an improvement of $1.05\%$ in Dice and $0.41$mm in HD95. Compared to the best-performing baseline, this was also a $0.44\%$ increase in Dice and $0.33$mm decrease in HD95. 
% Notably, we used an expansion factor $\alpha = 1$ in every Mamba block, so we spent $0.5$M less parameters than U-Mamba-Enc while using similar number of FLOPs.

\begin{table}[tb]
\centering
\renewcommand{\arraystretch}{1.2}
\caption{Comparisons on the ACDC test set. The best results are highlighted in bold.}
\label{tab:acdc_results}
\begin{tabular}{|l|r|r|c|c|c|c|}
\hline 
\multicolumn{1}{|c|}{\multirow{1}{*}{Method}} & \multicolumn{1}{c|}{\multirow{1}{*}{Size}} & \multicolumn{1}{c|}{\multirow{1}{*}{FLOPs}} & Dice (\%) $\uparrow$ & HD95 (\%) $\downarrow$ \\
\hline  % \midrule        
UNet3D~\cite{kerfoot2019left} & $5.7$M & $131.6$G & 90.92$_{\pm0.04}$ & 1.16$_{\pm0.01}$ \\
nnUNet~\cite{isensee2024nnu} & $191.8$M & $479.7$G & 91.88$_{\pm0.04}$ & 1.17$_{\pm0.07}$ \\
UNETR~\cite{hatamizadeh2022unetr} & $101.8$M & $240.5$G & 86.54$_{\pm0.09}$ & 2.49$_{\pm0.09}$ \\
Swin-UNETR~\cite{hatamizadeh2021swin} & $62.2$M & $769.4$G & 89.19$_{\pm0.19}$ & 2.17$_{\pm0.17}$ \\
NestedFormer~\cite{xing2022nestedformer} & $5.4$M & $201.7$G & 90.15$_{\pm0.11}$ & 1.90$_{\pm0.33}$ \\
EoFormer~\cite{she2023eoformer} & $6.4$M & $35.5$G & 91.12$_{\pm0.07}$ & 1.19$_{\pm0.02}$ \\
\hline
U-Mamba-Bot~\cite{ma2024u} & $57.3$M & $431.8$G & 91.94$_{\pm0.05}$ & 1.26$_{\pm0.15}$ \\
U-Mamba-Enc~\cite{ma2024u} & $59.9$M & $466.4$G & 91.65$_{\pm0.31}$ & 1.13$_{\pm0.02}$ \\
SegMamba~\cite{xing2024segmamba} & $66.9$M & $637.2$G & 90.74$_{\pm0.09}$ & 1.19$_{\pm0.02}$ \\
% \hline
\rowcolor{LightCyan}
Ours & $59.7$M & $467.9$G & \subbest{92.18$_{\pm0.13}$} & \subbest{1.10$_{\pm0.00}$} \\
% UNet3D~\cite{kerfoot2019left} & $5.7$M & $131.6$G & $90.92 \pm 0.03$ & $1.16 \pm 0.01$ \\
% nnUNet~\cite{isensee2024nnu} & $191.8$M & $479.7$G & $91.88 \pm 0.03$ & $1.17 \pm 0.05$ \\
% UNETR~\cite{hatamizadeh2022unetr} & $101.8$M & $240.5$G & $86.72 \pm 0.17$ & $2.35 \pm 0.10$ \\
% Swin-UNETR~\cite{hatamizadeh2021swin} & $62.2$M & $769.4$G & $89.53 \pm 0.14$ & $1.76 \pm 0.15$ \\
% U-Mamba-Bot~\cite{ma2024u} & $57.3$M & $431.8$G & $91.94 \pm 0.04$ & $1.26 \pm 0.11$ \\
% U-Mamba-Enc~\cite{ma2024u} & $59.9$M & $466.4$G & $91.76 \pm 0.23$ & $1.12 \pm 0.11$ \\
% SegMamba~\cite{xing2024segmamba} & $66.9$M & $637.2$G & $90.92 \pm 0.06$ & $1.15 \pm 0.02$ \\
% \hline
% Ours & $59.7$M & $461.2$G & $92.20 \pm 0.06$ & $1.10 \pm 0.01$ \\
\hline
\end{tabular}
\end{table}

\Paragraph{ACDC.} In~\cref{tab:acdc_results,tab:acdc_class_results}, we reported the experimental results on the ACDC dataset. Our network achieved a Dice score of $92.18\%$ and HD95 of $1.10$mm, which were $0.53\%$ higher in Dice and $0.03$mm lower in HD95 than the original U-Mamba-Enc. 
% The performance of our method was also the best among all methods on the ACDC dataset. 
In addition, \cref{fig:3d_comparison_acdc} demonstrates that our method consistently performed better than nnUNet, Swin-UNETER, and U-Mamba-Enc across all the fine-grained classes.
Compared to U-Mamba-Bot~\cite{ma2024u}, which reached the highest Dice score among the baselines, our method also performed better with improvements of $0.24\%$ in Dice and $0.16$mm in HD95. 
% While obtaining a significantly stronger results, our model used a similar level of network size and FLOPs as U-Mamba-Enc.

%---------------------------------------------------------------
\subsection{Ablation Studies}

We conducted ablation studies on one 2D dataset, Endoscopy, and one 3D dataset, BraTS. Firstly, we investigated the impact of LTX and GTX. Furthermore, we examined the effects of distinct approaches to concatenate global to local tokens, as well as employing multiple scanning directions.

\begin{table}[tb]
\centering    
\renewcommand{\arraystretch}{1.2}
\caption{Ablation studies on the Endoscopy and BraTS test sets. The best results are highlighted in bold, and the second-best ones are underlined.}
\begin{tabular}{|l|c|c|c|c|c|c|c|c|}
\hline 
\multicolumn{1}{|c|}{\multirow{2}{*}{Method}} & \multicolumn{3}{c}{Endoscopy (2D)} & \multicolumn{2}{|c|}{BraTS (3D)} \\
\cline{2-6}
 & Dice (\%) $\uparrow$ & IoU (\%) $\uparrow$ & NSD (\%) $\uparrow$ & Dice (\%) $\uparrow$ & HD95 (mm) $\downarrow$ \\
\hline
VSS & 71.23$_{\pm1.00}$ & 67.81$_{\pm0.99}$ & 72.77$_{\pm1.02}$ & 87.01$_{\pm0.10}$ & 4.38$_{\pm0.09}$ \\
G-VMamba & 72.64$_{\pm0.21}$ & 69.08$_{\pm0.22}$ & 74.24$_{\pm0.22}$ & \subsecond{87.99$_{\pm0.09}$} & \subsecond{4.05$_{\pm0.07}$} \\
L-VMamba & \subsecond{74.15$_{\pm0.12}$} & \subsecond{70.56$_{\pm0.13}$} & \subsecond{75.81$_{\pm0.13}$} & 87.71$_{\pm0.09}$ & 4.12$_{\pm0.03}$ \\
LoG-VMamba & \subbest{75.17$_{\pm0.24}$} & \subbest{71.68$_{\pm0.23}$} & \subbest{76.83$_{\pm0.25}$} & \subbest{88.06$_{\pm0.08}$} & \subbest{3.97$_{\pm0.04}$} \\
% baseline & $71.23 \pm 0.45$ & $67.81 \pm 0.44$ & $72.77 \pm 0.45$ & $86.90 \pm 0.05$ & $4.48 \pm 0.11$ \\
% D-Mamba & $72.64 \pm 0.21$ & $69.08 \pm 0.22$ & $74.24 \pm 0.22$ & $87.99 \pm 0.09$ & $4.05 \pm 0.07$ \\
% C-Mamba & $74.15 \pm 0.12$ & $70.56 \pm 0.13$ & $75.81 \pm 0.13$ & $87.71 \pm 0.09$ & $4.12 \pm 0.03$ \\
% DC-Mamba & $75.17 \pm 0.11$ & $71.68 \pm 0.10$ & $76.83 \pm 0.11$ & $88.03 \pm0.26$ & $3.97 \pm 0.04$ \\
\hline
\end{tabular}
\label{tab:ablation_components}
\end{table}

\Paragraph{Impact of Each Component.} In~\cref{tab:ablation_components}, instead of using the LoG-VMamba block, we experimented with using either L-VMamba or G-VMamba block in our 2D and 3D models. L-VMamba was more effective than the G-VMamba block on the 2D Endoscopy dataset. While G-VMamba led to an increase of $1.41\%$ in Dice and $1.47\%$ in NSD, L-VMamba resulted in improvements of $2.92\%$ in Dice and $3.04\%$ in NSD compared to the vanilla VSS. When compared to L-VMamba, the combined block LoG-VMamba outperformed with differences of $1.02\%$ in both Dice and NSD. On the other hand, G-VMamba was better than L-VMamba on the 3D BraTS dataset, outperforming by $0.28\%$ in Dice and $0.07$mm in HD95. The combination of both LTX and GTX resulted in further improvements of $0.07\%$ in Dice and $0.08$mm in HD95.

\Paragraph{Concatenation of Local and Global Tokens.} We used only G-VMamba in this experiment, and present the detailed results in~\cref{tab:ablation_placement}. The ``Interleaved'' strategy achieved the best performance on the Endoscopy dataset. As such, it helped G-VMamba to improve $0.28\%$ in Dice and $0.32\%$ in IoU in comparison with the second-best strategy ``Center''. The best-performing strategy on BraTS was not as obvious. While the ``Interleaved'' strategy reached the highest Dice at $87.99\%$, the lowest HD95 of $4.00$mm belonged to the ``Split'' strategy.

\Paragraph{Scanning Strategies.} We defined $M$ as the number of scanning directions and evaluated $3$ scanning strategies: $(M = 1)$ only horizontal scan was used; $(M = 2)$ both horizontal and vertical scan were used; $(M = 4)$ Mamba scanned horizontally and vertically in both forward and backward directions. 
% The default configuration was $M = 1$. 
As shown in~\cref{tab:ablation_directions}, we did not observe the benefits of employing multiple scanning directions in the SSM module for our models. On both datasets, using the most computationally efficient approach, $M = 1$, obtained the best performance. Utilizing more than one scanning direction negatively impacted the performance.
% On BraTS, using two scans led to a decrease in Dice of $0.18\%$ and an increase in HD95 of $0.23$mm.

\begin{table}[tb]
\centering    
\caption{Effects of using different concatenation strategies on the Endoscopy and BraTS test sets}
\label{tab:ablation_placement}
\begin{tabular}{|c|c|c|c|c|c|c|c|c|}
\hline 
\multicolumn{1}{|c|}{\multirow{2}{*}{Strategy}} & \multicolumn{3}{c|}{Endoscopy (2D)} & \multicolumn{2}{c|}{BraTS (3D)} \\
\cline{2-6}
 & Dice (\%) $\uparrow$ & IoU (\%) $\uparrow$ & NSD (\%) $\uparrow$ & Dice (\%) $\uparrow$ & HD95 (mm) $\downarrow$ \\
\hline
Head & $71.20_{\pm 0.16}$ & $67.63_{\pm 0.16}$ & $72.78_{\pm 0.17}$ & $87.71_{\pm 0.12}$ & $4.01_{\pm 0.02}$ \\
Split & $72.01_{\pm 0.21}$ & $68.42_{\pm 0.21}$ & $73.60_{\pm 0.21}$ & $87.72_{\pm 0.11}$ & $4.00_{\pm 0.04}$ \\
Center & $72.36_{\pm 0.30}$ & $68.76_{\pm 0.31}$ & $73.96_{\pm 0.31}$ & $87.51_{\pm 0.18}$ & $4.05_{\pm 0.05}$ \\
Interleaved & $72.64_{\pm 0.21}$ & $69.08_{\pm 0.22}$ & $74.24_{\pm 0.22}$ & $87.99_{\pm 0.09}$ & $4.05_{\pm 0.07}$ \\
% Head & $71.20 \pm 0.16$ & $67.63 \pm 0.16$ & $72.78 \pm 0.17$ & $87.71 \pm 0.12$ & $4.01 \pm 0.02$ \\
% Split & $72.01 \pm 0.21$ & $68.42 \pm 0.21$ & $73.60 \pm 0.21$ & $87.72 \pm 0.11$ & $4.00 \pm 0.04$ \\
% Center & $72.36 \pm 0.30$ & $68.76 \pm 0.31$ & $73.96 \pm 0.31$ & $87.51 \pm 0.18$ & $4.05 \pm 0.05$ \\
% Even & $72.64 \pm 0.21$ & $69.08 \pm 0.22$ & $74.24 \pm 0.22$ & $87.99 \pm 0.09$ & $4.05 \pm 0.07$ \\
\hline
\end{tabular}
\end{table}

\begin{table}[tb]
\centering    
\renewcommand{\arraystretch}{1.2}
\caption{Effects of using multiple scanning directions on performances on the Endoscopy and BraTS test sets. The substantially best results are highlighted in bold.}
\label{tab:ablation_directions}
\begin{tabular}{|p{0.5cm}|c|c|c|c|c|c|c|c|}
\hline 
\multicolumn{1}{|c|}{\multirow{2}{*}{$M$}} & \multicolumn{3}{c|}{Endoscopy (2D)} & \multicolumn{2}{c|}{BraTS (3D)} \\
\cline{2-6}
 & Dice (\%) $\uparrow$ & IoU (\%) $\uparrow$ & NSD (\%) $\uparrow$ & Dice (\%) $\uparrow$ & HD95 (mm) $\downarrow$ \\
\hline
\multicolumn{1}{|c|}{\multirow{1}{*}{$1$}} & \subbest{75.17$_{\pm0.24}$} & \subbest{71.68$_{\pm0.23}$} & \subbest{76.83$_{\pm0.25}$} & \subbest{88.06$_{\pm0.08}$} & \subbest{3.97$_{\pm0.04}$} \\
\multicolumn{1}{|c|}{\multirow{1}{*}{$2$}} & 74.03$_{\pm0.31}$ & 70.50$_{\pm0.31}$ & 75.67$_{\pm0.31}$ & 87.88$_{\pm0.08}$ & 4.20$_{\pm0.07}$ \\
\multicolumn{1}{|c|}{\multirow{1}{*}{$4$}} & 75.08$_{\pm0.13}$ & 71.59$_{\pm0.13}$ & 76.75$_{\pm0.13}$ & 87.84$_{\pm0.13}$ & 4.13$_{\pm0.13}$ \\
% $1$ & $75.17 \pm 0.11$ & $71.68 \pm 0.10$ & $76.83 \pm 0.11$ & $88.03 \pm0.26$ & $3.97 \pm 0.04$ \\
% $2$ & $74.03 \pm 0.31$ & $70.50 \pm 0.31$ & $75.67 \pm 0.31$ & $87.83 \pm 0.07$ & $3.90 \pm 0.05$ \\
% $4$ & $75.08 \pm 0.13$ & $71.59 \pm 0.13$ & $76.75 \pm 0.13$ & & \\
\hline
\end{tabular}
\end{table}

%===========================================================
\section{Conclusion}
In this study, we have introduced a straightforward, yet effective and efficient approach to advance the SSMs for MIS. Our proposed framework addresses the fundamental sequential limitations of SSM-based methods in handling high-dimensional data such as 2D and 3D medical images. As such, we propose the LTX and GTX modules to enhance tokens with both local and global receptive fields, which are inspired by
the strength of CNN and ViT respectively. We then leverage these two components to form the LoG-VMamba block.
% , based on VSS~\cite{liu2024vmamba}. 

% We propose two ideas to improve the modeling power of Vision Mamba. 
As our framework is not specifically designed for MIS, it may be applicable to other problems such as classification and detection, or even multimodal applications. It is, however, outside the scope of the present work, the focus of which is improving MIS. Our experiments show that LoG-VMamba can be well integrated into advanced segmentation models such as Swin-UMamba and U-Mamba-Enc, leading to consistent improvements across distinct 2D and 3D medical imaging datasets. 
Furthermore, our method's enriched tokens eliminate the need for a complex scanning strategy, thereby enhancing computational efficiency.

\clearpage

\bibliographystyle{splncs04}
\bibliography{main}

\begin{thebibliography}{10}
\providecommand{\url}[1]{\texttt{#1}}
\providecommand{\urlprefix}{URL }
\providecommand{\doi}[1]{https://doi.org/#1}

\bibitem{allan20192017}
Allan, M., Shvets, A., Kurmann, T., Zhang, Z., Duggal, R., Su, Y.H., Rieke, N., Laina, I., Kalavakonda, N., Bodenstedt, S., et~al.: 2017 robotic instrument segmentation challenge. arXiv preprint arXiv:1902.06426  (2019)

\bibitem{ba2016layer}
Ba, J.L., Kiros, J.R., Hinton, G.E.: Layer normalization. arXiv preprint arXiv:1607.06450  (2016)

\bibitem{bakas2017advancing}
Bakas, S., Akbari, H., Sotiras, A., Bilello, M., Rozycki, M., Kirby, J.S., Freymann, J.B., Farahani, K., Davatzikos, C.: Advancing the cancer genome atlas glioma mri collections with expert segmentation labels and radiomic features. Scientific data  \textbf{4}(1),  1--13 (2017)

\bibitem{bakas2018identifying}
Bakas, S., Reyes, M., Jakab, A., Bauer, S., Rempfler, M., Crimi, A., Shinohara, R.T., Berger, C., Ha, S.M., Rozycki, M., et~al.: Identifying the best machine learning algorithms for brain tumor segmentation, progression assessment, and overall survival prediction in the brats challenge. arXiv preprint arXiv:1811.02629  (2018)

\bibitem{bernard2018deep}
Bernard, O., Lalande, A., Zotti, C., Cervenansky, F., Yang, X., Heng, P.A., Cetin, I., Lekadir, K., Camara, O., Ballester, M.A.G., et~al.: Deep learning techniques for automatic mri cardiac multi-structures segmentation and diagnosis: is the problem solved? IEEE transactions on medical imaging  \textbf{37}(11),  2514--2525 (2018)

\bibitem{cao2022swin}
Cao, H., Wang, Y., Chen, J., Jiang, D., Zhang, X., Tian, Q., Wang, M.: Swin-unet: Unet-like pure transformer for medical image segmentation. In: European conference on computer vision. pp. 205--218. Springer (2022)

\bibitem{chollet2017xception}
Chollet, F.: Xception: Deep learning with depthwise separable convolutions. In: Proceedings of the IEEE conference on computer vision and pattern recognition. pp. 1251--1258 (2017)

\bibitem{dai2021coatnet}
Dai, Z., Liu, H., Le, Q.V., Tan, M.: Coatnet: Marrying convolution and attention for all data sizes. Advances in neural information processing systems  \textbf{34},  3965--3977 (2021)

\bibitem{dang2024singr}
Dang, T., Nguyen, H.H., Tiulpin, A.: Singr: Brain tumor segmentation via signed normalized geodesic transform regression. arXiv preprint arXiv:2405.16813  (2024)

\bibitem{dong2022cswin}
Dong, X., Bao, J., Chen, D., Zhang, W., Yu, N., Yuan, L., Chen, D., Guo, B.: Cswin transformer: A general vision transformer backbone with cross-shaped windows. In: Proceedings of the IEEE/CVF conference on computer vision and pattern recognition. pp. 12124--12134 (2022)

\bibitem{dosovitskiy2020image}
Dosovitskiy, A., Beyer, L., Kolesnikov, A., Weissenborn, D., Zhai, X., Unterthiner, T., Dehghani, M., Minderer, M., Heigold, G., Gelly, S., et~al.: An image is worth 16x16 words: Transformers for image recognition at scale. arXiv preprint arXiv:2010.11929  (2020)

\bibitem{gu2023mamba}
Gu, A., Dao, T.: Mamba: Linear-time sequence modeling with selective state spaces. arXiv preprint arXiv:2312.00752  (2023)

\bibitem{gu2020hippo}
Gu, A., Dao, T., Ermon, S., Rudra, A., R{\'e}, C.: Hippo: Recurrent memory with optimal polynomial projections. Advances in neural information processing systems  \textbf{33},  1474--1487 (2020)

\bibitem{gu2021efficiently}
Gu, A., Goel, K., R{\'e}, C.: Efficiently modeling long sequences with structured state spaces. arXiv preprint arXiv:2111.00396  (2021)

\bibitem{guo2022cmt}
Guo, J., Han, K., Wu, H., Xu, C., Tang, Y., Xu, C., Wang, Y.: Cmt: Convolutional neural networks meet vision transformers. 2022 ieee. In: CVF Conference on Computer Vision and Pattern Recognition (CVPR). pp. 12165--12175 (2022)

\bibitem{hassani2023neighborhood}
Hassani, A., Walton, S., Li, J., Li, S., Shi, H.: Neighborhood attention transformer. In: Proceedings of the IEEE/CVF Conference on Computer Vision and Pattern Recognition. pp. 6185--6194 (2023)

\bibitem{hatamizadeh2021swin}
Hatamizadeh, A., Nath, V., Tang, Y., Yang, D., Roth, H.R., Xu, D.: Swin unetr: Swin transformers for semantic segmentation of brain tumors in mri images. In: International MICCAI Brainlesion Workshop. pp. 272--284. Springer (2021)

\bibitem{hatamizadeh2022unetr}
Hatamizadeh, A., Tang, Y., Nath, V., Yang, D., Myronenko, A., Landman, B., Roth, H.R., Xu, D.: Unetr: Transformers for 3d medical image segmentation. In: Proceedings of the IEEE/CVF winter conference on applications of computer vision. pp. 574--584 (2022)

\bibitem{he2016deep}
He, K., Zhang, X., Ren, S., Sun, J.: Deep residual learning for image recognition. In: Proceedings of the IEEE conference on computer vision and pattern recognition. pp. 770--778 (2016)

\bibitem{howard2017mobilenets}
Howard, A.G., Zhu, M., Chen, B., Kalenichenko, D., Wang, W., Weyand, T., Andreetto, M., Adam, H.: Mobilenets: Efficient convolutional neural networks for mobile vision applications. arXiv preprint arXiv:1704.04861  (2017)

\bibitem{hu2018squeeze}
Hu, J., Shen, L., Sun, G.: Squeeze-and-excitation networks. In: Proceedings of the IEEE conference on computer vision and pattern recognition. pp. 7132--7141 (2018)

\bibitem{huang2017densely}
Huang, G., Liu, Z., Van Der~Maaten, L., Weinberger, K.Q.: Densely connected convolutional networks. In: Proceedings of the IEEE conference on computer vision and pattern recognition. pp. 4700--4708 (2017)

\bibitem{huang2023vision}
Huang, H., Zhou, X., Cao, J., He, R., Tan, T.: Vision transformer with super token sampling. In: Proceedings of the IEEE/CVF Conference on Computer Vision and Pattern Recognition. pp. 22690--22699 (2023)

\bibitem{huang2024localmamba}
Huang, T., Pei, X., You, S., Wang, F., Qian, C., Xu, C.: Localmamba: Visual state space model with windowed selective scan. arXiv preprint arXiv:2403.09338  (2024)

\bibitem{isensee2024nnu}
Isensee, F., Wald, T., Ulrich, C., Baumgartner, M., Roy, S., Maier-Hein, K., Jaeger, P.F.: nnu-net revisited: A call for rigorous validation in 3d medical image segmentation. arXiv preprint arXiv:2404.09556  (2024)

\bibitem{kalman1960new}
Kalman, R.E.: A new approach to linear filtering and prediction problems  (1960)

\bibitem{kerfoot2019left}
Kerfoot, E., Clough, J., Oksuz, I., Lee, J., King, A.P., Schnabel, J.A.: Left-ventricle quantification using residual u-net. In: Statistical Atlases and Computational Models of the Heart. Atrial Segmentation and LV Quantification Challenges: 9th International Workshop, STACOM 2018, Held in Conjunction with MICCAI 2018, Granada, Spain, September 16, 2018, Revised Selected Papers 9. pp. 371--380. Springer (2019)

\bibitem{kingma2014adam}
Kingma, D.P., Ba, J.: Adam: A method for stochastic optimization. arXiv preprint arXiv:1412.6980  (2014)

\bibitem{knigge2023modelling}
Knigge, D.M., Romero, D.W., Gu, A., Gavves, E., Bekkers, E.J., Tomczak, J.M., Hoogendoorn, M., jakob Sonke, J.: Modelling long range dependencies in \$n\$d: From task-specific to a general purpose {CNN}. In: The Eleventh International Conference on Learning Representations (2023), \url{https://openreview.net/forum?id=ZW5aK4yCRqU}

\bibitem{krizhevsky2017imagenet}
Krizhevsky, A., Sutskever, I., Hinton, G.E.: Imagenet classification with deep convolutional neural networks. Communications of the ACM  \textbf{60}(6),  84--90 (2017)

\bibitem{lecun2015deep}
LeCun, Y., Bengio, Y., Hinton, G.: Deep learning. Nature  \textbf{521}(7553),  436--444 (2015)

\bibitem{li2018h}
Li, X., Chen, H., Qi, X., Dou, Q., Fu, C.W., Heng, P.A.: H-denseunet: hybrid densely connected unet for liver and tumor segmentation from ct volumes. IEEE transactions on medical imaging  \textbf{37}(12),  2663--2674 (2018)

\bibitem{lin2023scale}
Lin, W., Wu, Z., Chen, J., Huang, J., Jin, L.: Scale-aware modulation meet transformer. In: Proceedings of the IEEE/CVF International Conference on Computer Vision. pp. 6015--6026 (2023)

\bibitem{liu2024swin}
Liu, J., Yang, H., Zhou, H.Y., Xi, Y., Yu, L., Yu, Y., Liang, Y., Shi, G., Zhang, S., Zheng, H., et~al.: Swin-umamba: Mamba-based unet with imagenet-based pretraining. arXiv preprint arXiv:2402.03302  (2024)

\bibitem{liu2024vmamba}
Liu, Y., Tian, Y., Zhao, Y., Yu, H., Xie, L., Wang, Y., Ye, Q., Liu, Y.: Vmamba: Visual state space model. arXiv preprint arXiv:2401.10166  (2024)

\bibitem{liu2021swin}
Liu, Z., Lin, Y., Cao, Y., Hu, H., Wei, Y., Zhang, Z., Lin, S., Guo, B.: Swin transformer: Hierarchical vision transformer using shifted windows. In: Proceedings of the IEEE/CVF international conference on computer vision. pp. 10012--10022 (2021)

\bibitem{liu2022convnet}
Liu, Z., Mao, H., Wu, C.Y., Feichtenhofer, C., Darrell, T., Xie, S.: A convnet for the 2020s. In: Proceedings of the IEEE/CVF conference on computer vision and pattern recognition. pp. 11976--11986 (2022)

\bibitem{ma2024u}
Ma, J., Li, F., Wang, B.: U-mamba: Enhancing long-range dependency for biomedical image segmentation. arXiv preprint arXiv:2401.04722  (2024)

\bibitem{ma2024multimodality}
Ma, J., Xie, R., Ayyadhury, S., Ge, C., Gupta, A., Gupta, R., Gu, S., Zhang, Y., Lee, G., Kim, J., et~al.: The multimodality cell segmentation challenge: toward universal solutions. Nature methods pp. 1--11 (2024)

\bibitem{menze2014multimodal}
Menze, B.H., Jakab, A., Bauer, S., Kalpathy-Cramer, J., Farahani, K., Kirby, J., Burren, Y., Porz, N., Slotboom, J., Wiest, R., et~al.: The multimodal brain tumor image segmentation benchmark (brats). IEEE transactions on medical imaging  \textbf{34}(10),  1993--2024 (2014)

\bibitem{orvieto2023resurrecting}
Orvieto, A., Smith, S.L., Gu, A., Fernando, A., Gulcehre, C., Pascanu, R., De, S.: Resurrecting recurrent neural networks for long sequences. In: International Conference on Machine Learning. pp. 26670--26698. PMLR (2023)

\bibitem{pan2022integration}
Pan, X., Ge, C., Lu, R., Song, S., Chen, G., Huang, Z., Huang, G.: On the integration of self-attention and convolution. In: Proceedings of the IEEE/CVF conference on computer vision and pattern recognition. pp. 815--825 (2022)

\bibitem{paszke2019pytorch}
Paszke, A., Gross, S., Massa, F., Lerer, A., Bradbury, J., Chanan, G., Killeen, T., Lin, Z., Gimelshein, N., Antiga, L., et~al.: Pytorch: An imperative style, high-performance deep learning library. Advances in neural information processing systems  \textbf{32} (2019)

\bibitem{ronneberger2015u}
Ronneberger, O., Fischer, P., Brox, T.: U-net: Convolutional networks for biomedical image segmentation. In: Medical image computing and computer-assisted intervention--MICCAI 2015: 18th international conference, Munich, Germany, October 5-9, 2015, proceedings, part III 18. pp. 234--241. Springer (2015)

\bibitem{she2023eoformer}
She, D., Zhang, Y., Zhang, Z., Li, H., Yan, Z., Sun, X.: Eoformer: Edge-oriented transformer for brain tumor segmentation. In: International Conference on Medical Image Computing and Computer-Assisted Intervention. pp. 333--343. Springer (2023)

\bibitem{shi2024multi}
Shi, Y., Dong, M., Xu, C.: Multi-scale vmamba: Hierarchy in hierarchy visual state space model. arXiv preprint arXiv:2405.14174  (2024)

\bibitem{simonyan2014very}
Simonyan, K., Zisserman, A.: Very deep convolutional networks for large-scale image recognition. arXiv preprint arXiv:1409.1556  (2014)

\bibitem{touvron2021training}
Touvron, H., Cord, M., Douze, M., Massa, F., Sablayrolles, A., J{\'e}gou, H.: Training data-efficient image transformers \& distillation through attention. In: International conference on machine learning. pp. 10347--10357. PMLR (2021)

\bibitem{vaswani2017attention}
Vaswani, A., Shazeer, N., Parmar, N., Uszkoreit, J., Jones, L., Gomez, A.N., Kaiser, {\L}., Polosukhin, I.: Attention is all you need. Advances in neural information processing systems  \textbf{30} (2017)

\bibitem{wang2024mamba}
Wang, F., Wang, J., Ren, S., Wei, G., Mei, J., Shao, W., Zhou, Y., Yuille, A., Xie, C.: Mamba-r: Vision mamba also needs registers. arXiv preprint arXiv:2405.14858  (2024)

\bibitem{xiao2018weighted}
Xiao, X., Lian, S., Luo, Z., Li, S.: Weighted res-unet for high-quality retina vessel segmentation. In: 2018 9th international conference on information technology in medicine and education (ITME). pp. 327--331. IEEE (2018)

\bibitem{xing2024segmamba}
Xing, Z., Ye, T., Yang, Y., Liu, G., Zhu, L.: Segmamba: Long-range sequential modeling mamba for 3d medical image segmentation. arXiv preprint arXiv:2401.13560  (2024)

\bibitem{xing2022nestedformer}
Xing, Z., Yu, L., Wan, L., Han, T., Zhu, L.: Nestedformer: Nested modality-aware transformer for brain tumor segmentation. In: International Conference on Medical Image Computing and Computer-Assisted Intervention. pp. 140--150. Springer (2022)

\bibitem{yu2015multi}
Yu, F., Koltun, V.: Multi-scale context aggregation by dilated convolutions. arXiv preprint arXiv:1511.07122  (2015)

\bibitem{yu2022metaformer}
Yu, W., Luo, M., Zhou, P., Si, C., Zhou, Y., Wang, X., Feng, J., Yan, S.: Metaformer is actually what you need for vision. In: Proceedings of the IEEE/CVF conference on computer vision and pattern recognition. pp. 10819--10829 (2022)

\bibitem{zhao2024rs}
Zhao, S., Chen, H., Zhang, X., Xiao, P., Bai, L., Ouyang, W.: Rs-mamba for large remote sensing image dense prediction. arXiv preprint arXiv:2404.02668  (2024)

\bibitem{zhou2021deepvit}
Zhou, D., Kang, B., Jin, X., Yang, L., Lian, X., Jiang, Z., Hou, Q., Feng, J.: Deepvit: Towards deeper vision transformer. arXiv preprint arXiv:2103.11886  (2021)

\bibitem{zhou2019unet++}
Zhou, Z., Siddiquee, M.M.R., Tajbakhsh, N., Liang, J.: Unet++: Redesigning skip connections to exploit multiscale features in image segmentation. IEEE transactions on medical imaging  \textbf{39}(6),  1856--1867 (2019)

\bibitem{zhu2024vision}
Zhu, L., Liao, B., Zhang, Q., Wang, X., Liu, W., Wang, X.: Vision mamba: Efficient visual representation learning with bidirectional state space model. arXiv preprint arXiv:2401.09417  (2024)

\end{thebibliography}

\clearpage

\renewcommand{\thepage}{S\arabic{page}} 
\renewcommand{\thesection}{S\arabic{section}}  
\renewcommand{\thetable}{S\arabic{table}}  
\renewcommand{\thefigure}{S\arabic{figure}}

\setcounter{page}{1}
\setcounter{figure}{0}
\setcounter{table}{0}

\section*{Supplementary Material}

% \begin{figure}[ht]
%   \centering
%   \croppdf{figures/dc_mamba_block_vss}
%   \croppdf{figures/dc_mamba_block_dm}
%   \croppdf{figures/dc_mamba_block_cm}
%   \croppdf{figures/dc_mamba_block_dcm}
%   % \hspace*{\fill}
%   \subfloat[VSS block\label{fig:vss}]{
%   \includegraphics[height=0.42\textwidth]{figures/dc_mamba_block_vss-crop.pdf}}
%   \hfill  
%   \subfloat[C-Mamba block]{\includegraphics[height=0.42\textwidth]{figures/dc_mamba_block_cm-crop.pdf}}
%   \hfill
%   \subfloat[D-Mamba block]{\includegraphics[height=0.42\textwidth]{figures/dc_mamba_block_dm-crop.pdf}}
%   \hfill
%   \subfloat[DC-Mamba block]{\includegraphics[height=0.42\textwidth]{figures/dc_mamba_block_dcm-crop.pdf}}
%   % \hspace*{\fill}
%   \caption{The structural comparison between the VSS block~\cite{liu2024vmamba} and our proposed modules. White blocks indicate modules without learnable parameters. $S$ denotes the squeeze factor and $f$ is the downsampling factor. $C^\prime = \frac{CR^2}{S}$ is the feature size of S6 in Convolutional Mamba, where $R$ is the window size.}
%   \label{fig:mamba_variations}
% \end{figure}

\begin{figure}[ht]
  \centering
  \hspace*{\fill}
  \subfloat[2D segmentation model]{\includegraphics[height=0.6\textwidth,valign=t]{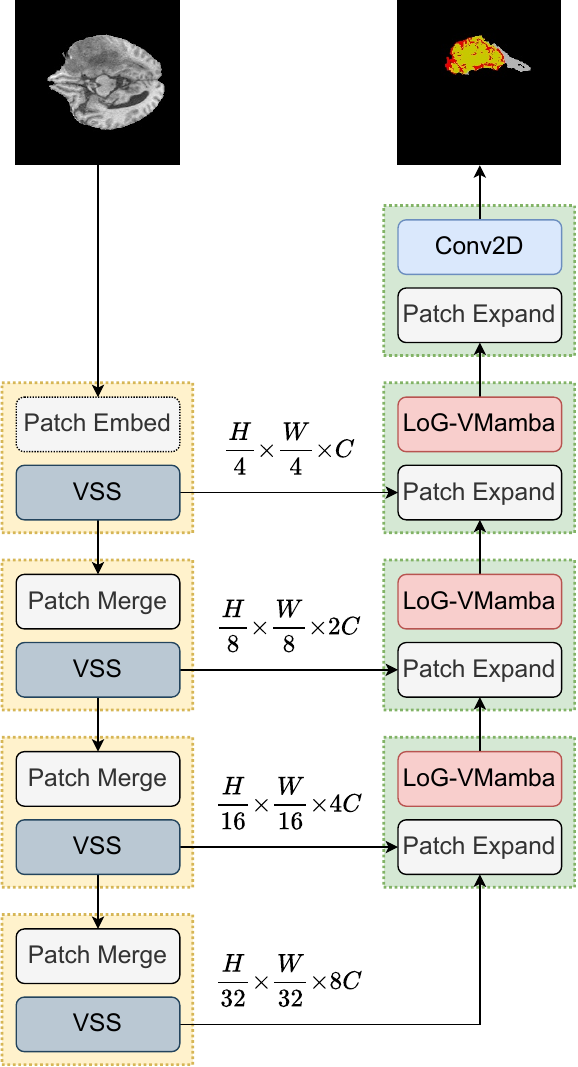}}
  \hfill
  \subfloat[3D segmentation model]{\includegraphics[height=0.6\textwidth,valign=t]{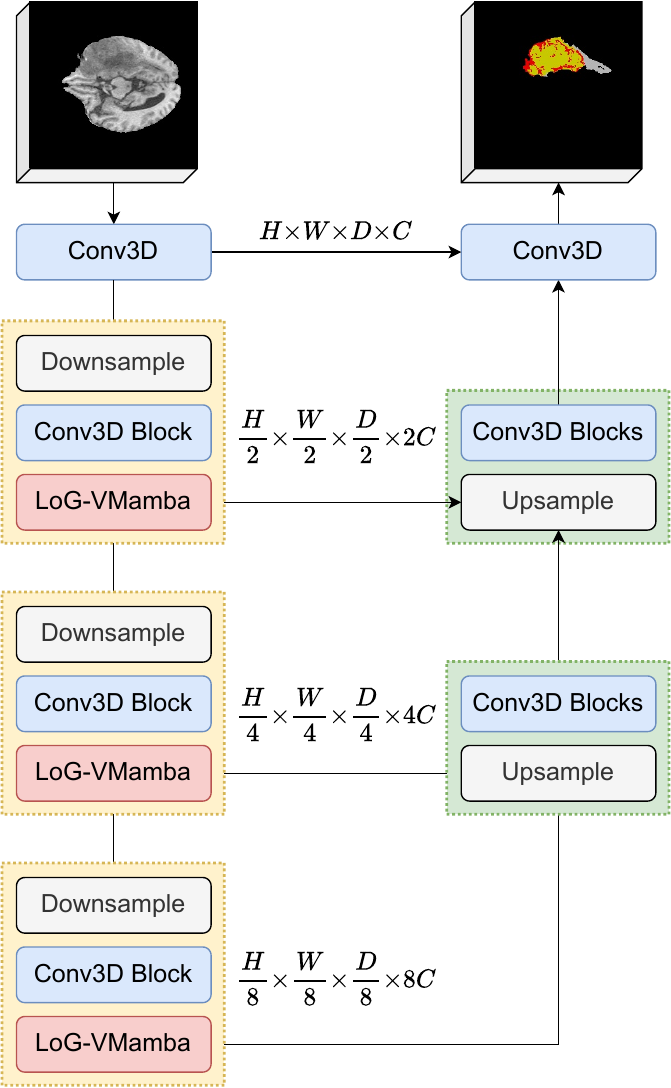}}
  \hspace*{\fill}
  \caption{The overview of our 2D and 3D segmentation models. These illustrations serve as a conceptual representation. Following~\cite{liu2024swin,ma2024u}, the number of blocks differs across datasets. The detailed configurations are shown in~\cref{tab:configs}.}
  \label{fig:dc_segmentation}
\end{figure}

\begin{table}[ht]
\centering    
\renewcommand{\arraystretch}{1.4}
\caption{Performance comparisons on the BraTS test set for each class. The best results are highlighted in bold while the second-best ones are underlined.}
\label{tab:brats_class_results}
\begin{tabular}{|l|c|c|c|c|c|c|c|c|}
\hline 
\multicolumn{1}{|c|}{\multirow{2}{*}{Method}} & \multicolumn{4}{c|}{Dice score (\%) $\uparrow$} & \multicolumn{4}{c|}{HD95 ($mm$) $\downarrow$} \\
\cline{2-9}
 & ET & TC & WT & Avg & ET & TC & WT & Avg \\
\hline
% TransBTS~\cite{wang2021transbts} & 81.0$_{\pm0.3}$ & 83.0$_{\pm0.4}$ & 90.4$_{\pm0.1}$ & 84.8$_{\pm0.2}$ & 4.2$_{\pm0.3}$ & 6.4$_{\pm0.2}$ & 6.4$_{\pm0.1}$ & 5.7$_{\pm0.2}$ \\
% SegResnet~\cite{myronenko20193d} & 81.1$_{\pm0.3}$ & 85.5$_{\pm0.3}$ & 90.7$_{\pm0.1}$ & 85.8$_{\pm0.2}$ & 3.3$_{\pm0.1}$ & 5.5$_{\pm0.3}$ & 5.7$_{\pm0.4}$ & 4.9$_{\pm0.2}$ \\
UNet3D~\cite{kerfoot2019left} & 83.1$_{\pm0.2}$ & 86.1$_{\pm0.3}$ & 90.4$_{\pm0.2}$ & 86.5$_{\pm0.1}$ & 3.8$_{\pm0.4}$ & 5.9$_{\pm0.3}$ & 6.5$_{\pm0.6}$ & 5.4$_{\pm0.4}$ \\
nnUNet~\cite{isensee2024nnu} & 84.1$_{\pm0.3}$ & \underline{87.2$_{\pm0.5}$} & 91.3$_{\pm0.1}$ & 87.5$_{\pm0.3}$ & 3.3$_{\pm0.5}$ & 5.7$_{\pm0.6}$ & 5.5$_{\pm0.3}$ & 4.8$_{\pm0.4}$ \\
UNETR~\cite{hatamizadeh2022unetr} & 82.3$_{\pm0.2}$ & 82.0$_{\pm0.4}$ & 90.0$_{\pm0.1}$ & 84.8$_{\pm0.2}$ & 4.0$_{\pm0.3}$ & 7.4$_{\pm0.2}$ & 6.2$_{\pm0.4}$ & 5.9$_{\pm0.2}$ \\
Swin-UNETR~\cite{hatamizadeh2021swin} & 84.1$_{\pm0.2}$ & 85.7$_{\pm0.4}$ & 90.9$_{\pm0.1}$ & 86.9$_{\pm0.2}$ & 3.7$_{\pm0.3}$ & 6.4$_{\pm0.2}$ & 6.0$_{\pm0.2}$ & 5.4$_{\pm0.2}$ \\
NestedFormer~\cite{xing2022nestedformer} & 83.5$_{\pm0.1}$ & 85.4$_{\pm0.1}$ & 91.2$_{\pm0.1}$ & 86.7$_{\pm0.1}$ & 4.4$_{\pm0.4}$ & 7.4$_{\pm0.4}$ & 6.6$_{\pm0.4}$ & 6.1$_{\pm0.4}$ \\
EoFormer~\cite{she2023eoformer} & 82.5$_{\pm0.2}$ & 84.8$_{\pm0.4}$ & 91.3$_{\pm0.0}$ & 86.2$_{\pm0.2}$ & 3.7$_{\pm0.2}$ & 6.4$_{\pm0.4}$ & 5.9$_{\pm0.3}$ & 5.3$_{\pm0.2}$ \\
\hline
U-Mamba-Bot~\cite{ma2024u} & 84.1$_{\pm0.2}$ & 87.0$_{\pm0.3}$ & \underline{91.5$_{\pm0.1}$} & 87.5$_{\pm0.2}$ & \underline{2.9$_{\pm0.1}$} & \underline{5.0$_{\pm0.3}$} & \subbest{5.0$_{\pm0.2}$} & \underline{4.3$_{\pm0.2}$} \\
U-Mamba-Enc~\cite{ma2024u} & 83.7$_{\pm0.1}$ & 86.2$_{\pm0.2}$ & 91.2$_{\pm0.1}$ & 87.0$_{\pm0.1}$ & 3.1$_{\pm0.2}$ & \underline{5.0$_{\pm0.2}$} & \underline{5.1$_{\pm0.1}$} & 4.4$_{\pm0.1}$ \\
SegMamba~\cite{xing2024segmamba} & \subbest{85.0$_{\pm0.2}$} & 86.7$_{\pm0.3}$ & 91.2$_{\pm0.1}$ & \underline{87.6$_{\pm0.2}$} & 3.2$_{\pm0.3}$ & 5.8$_{\pm0.4}$ & 5.2$_{\pm0.1}$ & 4.7$_{\pm0.2}$ \\
% \hline
\rowcolor{LightCyan}
Ours & \underline{84.7$_{\pm0.2}$} & \subbest{87.9$_{\pm0.3}$} & \subbest{91.6$_{\pm0.1}$} & \subbest{88.1$_{\pm0.1}$} & \subbest{2.4$_{\pm0.1}$} & \subbest{4.5$_{\pm0.1}$} & \subbest{5.0$_{\pm0.2}$} & \subbest{4.0$_{\pm0.0}$} \\
\hline
\end{tabular}
\end{table}

\begin{table}[ht]
\centering
\renewcommand{\arraystretch}{1.4}
\caption{Performance comparisons on the ACDC test set for each class. The best results are highlighted in bold while the second-best ones are underlined.}
\label{tab:acdc_class_results}
\begin{tabular}{|l|c|c|c|c|c|c|c|c|}
\hline 
\multicolumn{1}{|c|}{\multirow{2}{*}{Method}} & \multicolumn{4}{c|}{Dice score (\%) $\uparrow$} & \multicolumn{4}{c|}{HD95 ($mm$) $\downarrow$} \\
\cline{2-9}
 & RV & MYO & LV & Avg & RV & Myo & LV & Avg \\
\hline
UNet3D~\cite{kerfoot2019left} & 90.2$_{\pm0.1}$ & 89.3$_{\pm0.1}$ & 93.3$_{\pm0.1}$ & 90.9$_{\pm0.0}$ & \underline{1.3$_{\pm0.0}$} & \underline{1.1$_{\pm0.0}$} & \underline{1.2$_{\pm0.0}$} & \underline{1.2$_{\pm0.0}$} \\
nnUNet~\cite{isensee2024nnu} & 91.4$_{\pm0.0}$ & 89.9$_{\pm0.0}$ & \subbest{94.3$_{\pm0.1}$} & \underline{91.9$_{\pm0.0}$} & \subbest{1.2$_{\pm0.0}$} & \subbest{1.0$_{\pm0.0}$} & 1.3$_{\pm0.2}$ & \underline{1.2$_{\pm0.1}$} \\
UNETR~\cite{hatamizadeh2022unetr} & 85.0$_{\pm0.2}$ & 84.7$_{\pm0.2}$ & 89.9$_{\pm0.2}$ & 86.5$_{\pm0.1}$ & 2.8$_{\pm0.1}$ & 2.1$_{\pm0.1}$ & 2.6$_{\pm0.1}$ & 2.5$_{\pm0.1}$ \\
Swin-UNETR~\cite{hatamizadeh2021swin} & 87.9$_{\pm0.2}$ & 87.5$_{\pm0.2}$ & 92.1$_{\pm0.3}$ & 89.2$_{\pm0.2}$ & 2.8$_{\pm0.3}$ & 1.4$_{\pm0.1}$ & 2.3$_{\pm0.2}$ & 2.2$_{\pm0.2}$ \\
NestedFormer~\cite{xing2022nestedformer} & 89.2$_{\pm0.1}$ & 88.3$_{\pm0.1}$ & 92.9$_{\pm0.1}$ & 90.1$_{\pm0.1}$ & 1.6$_{\pm0.2}$ & 1.6$_{\pm0.4}$ & 2.5$_{\pm0.6}$ & 1.9$_{\pm0.3}$ \\
EoFormer~\cite{she2023eoformer} & 89.9$_{\pm0.0}$ & 89.8$_{\pm0.0}$ & 93.7$_{\pm0.2}$ & 91.1$_{\pm0.1}$ & \underline{1.3$_{\pm0.0}$} & \subbest{1.0$_{\pm0.0}$} & \underline{1.2$_{\pm0.1}$} & \underline{1.2$_{\pm0.0}$} \\
\hline
U-Mamba-Bot~\cite{ma2024u} & \underline{91.6$_{\pm0.1}$} & \underline{90.2$_{\pm0.0}$} & 94.0$_{\pm0.1}$ & \underline{91.9$_{\pm0.1}$} & \subbest{1.2$_{\pm0.0}$} & 1.3$_{\pm0.2}$ & 1.4$_{\pm0.2}$ & 1.3$_{\pm0.1}$ \\
U-Mamba-Enc~\cite{ma2024u} & 91.1$_{\pm0.4}$ & 89.9$_{\pm0.3}$ & 93.9$_{\pm0.2}$ & 91.6$_{\pm0.3}$ & \subbest{1.2$_{\pm0.0}$} & \subbest{1.0$_{\pm0.0}$} & \subbest{1.1$_{\pm0.0}$} & \subbest{1.1$_{\pm0.0}$} \\
SegMamba~\cite{xing2024segmamba} & 89.6$_{\pm0.2}$ & 89.0$_{\pm0.1}$ & 93.6$_{\pm0.0}$ & 90.7$_{\pm0.1}$ & \underline{1.3$_{\pm0.0}$} & \underline{1.1$_{\pm0.0}$} & \underline{1.2$_{\pm0.1}$} & \underline{1.2$_{\pm0.0}$} \\
% \hline
\rowcolor{LightCyan}
Ours & \subbest{92.0$_{\pm0.1}$} & \subbest{90.3$_{\pm0.0}$} & \underline{94.2$_{\pm0.1}$} & \subbest{92.2$_{\pm0.0}$} & \subbest{1.2$_{\pm0.0}$} & \subbest{1.0$_{\pm0.0}$} & \subbest{1.1$_{\pm0.0}$} & \subbest{1.1$_{\pm0.0}$} \\
\hline
\end{tabular}
\end{table}

\end{document}